\documentclass[letterpaper, 10 pt, conference]{ieeeconf}
\usepackage{times}
\usepackage[pdftex]{graphicx}
\usepackage{subfigure}
\usepackage{amsmath,amssymb,amsopn,amstext,amsfonts}
\usepackage{cancel}
\usepackage[space]{cite}
\usepackage{pdfsync}
\usepackage{balance}
\usepackage{color}
\usepackage{mathtools}
\usepackage{pdfpages}
\usepackage{multirow}
\usepackage{makecell}
\usepackage{algorithmicx}  
\usepackage{algpseudocode} 
\usepackage{bm}

\usepackage{diagbox}
\usepackage{float}
\usepackage{epstopdf}
\usepackage{pifont}
\usepackage{multirow}
\usepackage{url}
\usepackage{tabularx}
\usepackage[linkcolor=black,citecolor=black,urlcolor=black,colorlinks=true]{hyperref}
\usepackage{verbatim} 
\usepackage[skip=3pt, font={small}]{caption}
\usepackage[linesnumbered,ruled,vlined]{algorithm2e}
\newcommand{\changed}[1]{\textcolor{black}{#1}}
\usepackage{tikz}
\usepackage{etoolbox}
\newcommand{\circled}[2][]{\tikz[baseline=(char.base)]
	{\node[shape = circle, draw, inner sep = 1pt]
		(char) {\phantom{\ifblank{#1}{#2}{#1}}};%
		\node at (char.center) {\makebox[0pt][c]{#2}};}}
\robustify{\circled}

\SetKwInput{KwInput}{Input}                
\SetKwInput{KwOutput}{Output}              

\graphicspath{{../Fig/}}
\DeclareGraphicsExtensions{.png,.jpg,.eps,.pdf}
\IEEEoverridecommandlockouts
\title{\LARGE \bf Autonomous Exploration with Terrestrial-Aerial Bimodal Vehicles}
\author{Yuman Gao$^*$, Ruibin Zhang$^*$, Tiancheng Lai$^*$, Yanjun Cao, Chao Xu, and Fei Gao$^{\dagger}$
	\thanks{ $^*$\textbf{Equal contribution.}
	All authors are with the Institute of Cyber-Systems and Control, College of Control Science and Engineering, Zhejiang University, Hangzhou 310027, China, and also with the Huzhou Institute, Zhejiang University, Huzhou 313000, China.
	}
        \thanks{$^\dagger$Corresponding author: Fei Gao.}
        \thanks{Email: {\tt\small \{ymgao, fgaoaa\}@zju.edu.cn}.}
}

\makeatletter
\let\@oldmaketitle\@maketitle

\makeatother

\begin{document}

\maketitle

\begin{abstract}
\label{sec:abstract}
Terrestrial-aerial bimodal vehicles, which integrate the high mobility of aerial robots with the long endurance of ground robots, offer significant potential for autonomous exploration.
Given the inherent energy and time constraints in practical exploration tasks, we present a hierarchical framework for the bimodal vehicle to utilize its flexible locomotion modalities for exploration.
Beginning with extracting environmental information to identify informative regions, we generate a set of potential bimodal viewpoints.
To adaptively manage energy and time constraints, we introduce an extended Monte Carlo Tree Search approach that strategically optimizes both modality selection and viewpoint sequencing.
Combined with an improved bimodal vehicle motion planner, we present a complete bimodal energy- and time-aware exploration system.
Extensive simulations and deployment on a customized real-world platform demonstrate the effectiveness of our system.
\end{abstract}
\section{Introduction}
\label{sec:introduction}

	Autonomous exploration has gained increasing attention in both academia and industry, with applications in search and rescue, engineering surveying, and tunnel inspection.
	In recent years, researchers have proposed considerable exploration strategies and deployed them on unmanned aerial vehicles (UAVs) and unmanned ground vehicles (UGVs). 
    However, exploration performance is constrained by the kinodynamic characteristics of mobile robots.
    Although aerial robots offer high mobility and a broad field of view (FoV), their endurance is significantly shorter than that of ground robots, limiting their ability to support large-scale and long-term exploration.
	Ground robots, especially wheeled vehicles, face challenges when navigating complex and rugged terrains, restricting exploration to wide, flat areas.
	To break the above hardware constraints, collaborative aerial-ground exploration systems consisting of UAV-UGV robots are proposed \cite{butzke20153, wang2018crash, ropero2019terra, qin2019autonomous, williams2020online}. However, adopting such systems introduces issues of multi-robot SLAM, planning, and coordination, which greatly increase the complexity of the problem.
	
\begin{figure}[t]
	\begin{center}
		\includegraphics[width=0.9\columnwidth]{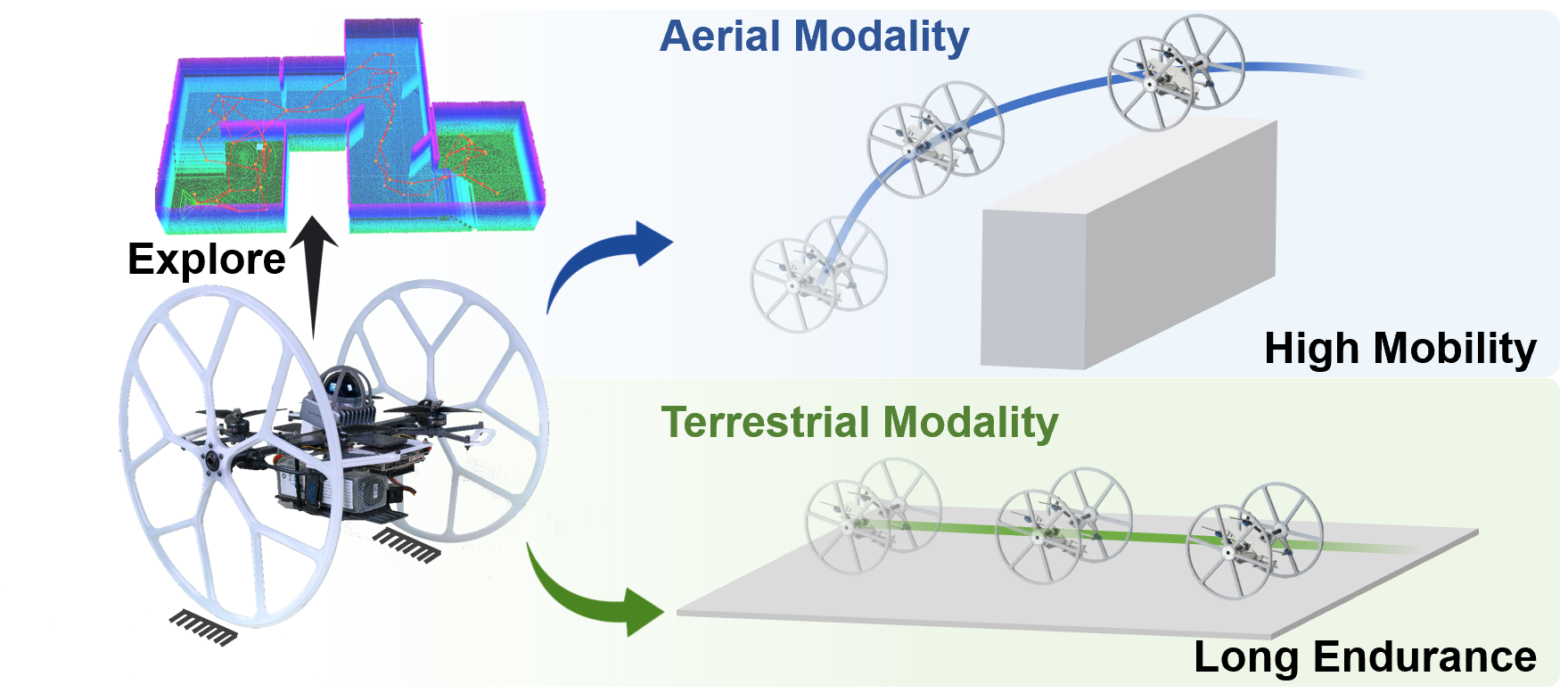}
	\end{center}
	\vspace{-0.2cm}
	\caption{
		\label{fig:head} The TABV integrates two modalities into a single platform, offering significant potential for autonomous exploration.
	}
	\vspace{-1.4cm}
\end{figure}

	
	To address the above issues, we propose a hierarchical exploration framework making use of a terrestrial-aerial bimodal vehicle (TABV)\cite{zhang2022tie}. With this type of vehicle, the long endurance of UGVs and high mobility and broad FoV of UAVs can be integrated into a single robot system as \changed{shown in} Fig.~\ref{fig:head}, showcasing great potential in exploration. 
    \changed{Moreover, in exploration tasks, especially in search and rescue scenarios, energy and time constraints should be considered, as robots operate with finite battery capacity and are typically expected to complete missions within a reasonable time frame.}
	However, these constraints are often overlooked in previous studies despite their significance. 
	With bimodal capability, TABV exhibits enhanced flexibility for these constraints.
	To leverage the unique characteristics of the TABV for exploration, we start by generating bimodal viewpoints according to the frontier of the known environment. 
	Subsequently, we introduce an adaptive exploration planner that enables the TABV to select a suitable modality to complete the exploration under given energy and time constraints. 
	We propose the Bimodal Monte Carlo Tree Search (BM-MCTS) method to determine the traverse sequence of the generated viewpoints. 
    Then, we adopt and improve the bimodal motion planner from our previous work \cite{zhang2022tie} for trajectory generation.
	To demonstrate and validate the proposed method, we conduct extensive exploration tests in various scenes in the simulation. Furthermore, we deploy our system into a customized TABV platform to conduct real-world experiments.
	


Contributions of this paper are summarized as follows:
\begin{enumerate}
    
    \item A hierarchical exploration framework for TABV, \changed{ featuring a bimodal viewpoint generation module based on two alternative coverage strategies, and an energy- and time-aware decision-making mechanism that fully exploits the robot’s bimodal locomotion capability.}
    
    \item \changed{An adaptive BM-MCTS approach for information-driven exploration, enabling} flexible modality and viewpoint selection under energy and time constraints.
	

    \item Integrating the exploration planner with an enhanced bimodal motion planner, \changed{featuring terrain perception and modality-aware planning}, forming a complete autonomous TABV system deployed on the real platform.
	
\end{enumerate}

\begin{figure*}[t]
	\begin{center}
		\includegraphics[width=2.0\columnwidth]{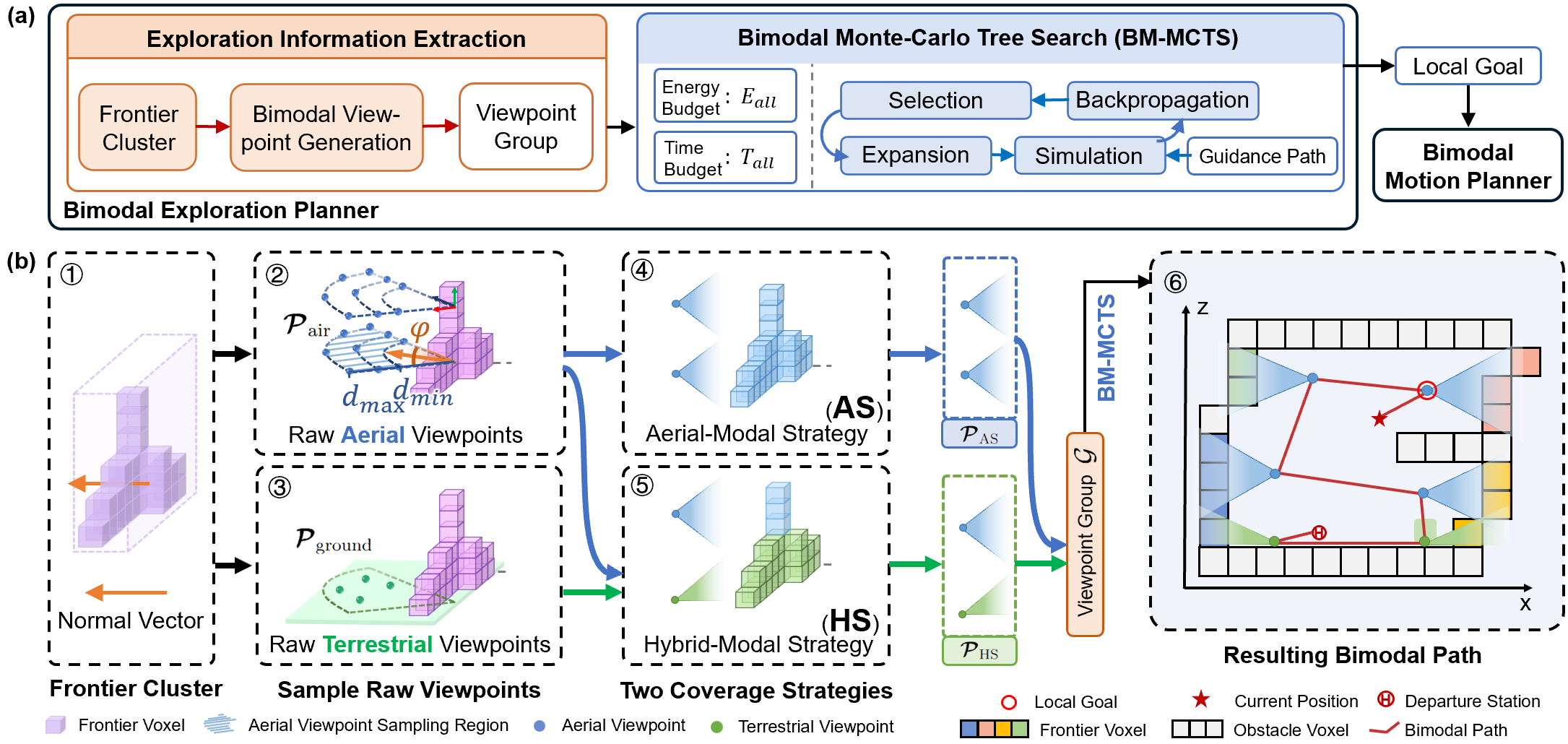}
	\end{center}
	\vspace{-0.3cm}
	\caption{
		\label{fig:cluster} 
        (a): An overview of the proposed TABV exploration framework. 
        (b): Module details: 
        \circled[1]{1}: \changed{A frontier cluster $C$ whose surface normal is computed via Principal Component Analysis (PCA) and oriented toward the known space.
		\circled[1]{2}-\circled[1]{3}: Raw aerial viewpoints $\bm{\mathcal{P}}_\text{aerial}$ and raw terrestrail viewpoints $\bm{\mathcal{P}}_\text{ground}$ generation. $\bm{\mathcal{P}}_\text{aerial}$ are generated via cylindrical coordinate sampling within $[d_{min},d_{max}]$ range and $\varphi$ angle around the cluster’s normal at multiple heights.
        \circled[1]{4}-\circled[1]{5}: Two alternative strategies to select bimodal viewpoints to fully cover $C$, resulting $\bm{\mathcal{P}}_\text{AS}$ and $\bm{\mathcal{P}}_\text{HS}$.
		\circled[1]{6}: An example of the resulting bimodal path and the next local goal.}
	}
	\vspace{-0.6cm}
\end{figure*}

\section{Related Work}
\label{sec:related work}

Autonomous exploration problem has been tackled using various strategies on multiple robot platforms. Among the various exploration methods, frontier-based methods make unknown environment information implied in frontiers to guide exploration\cite{yamauchi1997frontier, cieslewski2017rapid, cao2021tare,zhou2021fuel}.
Taking a set of frontiers or sampled viewpoints as goals, a global path traversing the goals is found through different approaches, such as the shortest distance criterion\cite{yamauchi1997frontier}, minimum velocity change criterion\cite{cieslewski2017rapid}, the balanced reward between information gain and path cost \cite{bircher2016receding}, or the solution of Traveling Salesman Problem (TSP)\cite{cao2021tare,zhou2021fuel}. 
Meanwhile, some works utilize MCTS to find non-myopic solutions for more complex such as decentralized, long-horizon, and multi-agent exploration tasks \cite{best2019dec, seiler2024multi, bone2023decentralised}.

To combine the advantages of aerial and ground mobile robots, researchers also focus on developing collaborative UAV-UGV exploration systems. Butzke et al.\cite{butzke20153} mount a drone on a ground robot as a backup. When the ground robot encounters high, invisible areas, the drone takes off to cover them. Wang et al.\cite{wang2018crash} \changed{use a centralized approach to plan UAV-UGV exploration trajectories}. Ground robots are preferable for open areas, while aerial robots are preferable for cluttered environments. Ropero et al.\cite{ropero2019terra} propose a path-planning algorithm for cooperative UGV–UAV exploration. Their strategy employs the ground robot as a mobile charging station to address the aerial robot's energy constraints, while the aerial robot reaches target points to overcome the ground robot's functionality limits. However, these multi-robot systems introduce increased system-level complexity, making deployment more challenging.

Not only the joint planning problem and communication problem between two platforms need to be solved, but the multi-robot SLAM problem also has to be concerned for an integrated system. To achieve scalability to multi-robot SLAM loop closures, Team CSIRO \cite{williams2020online} directly models the frontiers on point clouds to avoid dense volumetric map representations. Qin et al. \cite{qin2019autonomous} implement a two-layered exploration strategy, where the ground robot generates a coarse environment model, and the aerial robot produces the 3D fine mapping according to the coarse map.

Combining the advantages of ground and aerial robots without introducing the complexity of multiple platforms, TABVs have become a research hotspot \cite{10759756, 10341984, 10538378, tang2025duawlfin}. However, the works that use TABVs for exploration are rare. Rollocopter, the TABV of the team CoSTAR has been deployed for DARPA’s Subterranean Challenge. However, as just one among dozens of platforms, they utilize a general exploration planner without specifically addressing the schedule of the bimodal vehicle \cite{morrell2022nebula}.

In this work, we extend the MCTS method to address the bimodal, energy- and time-constrained exploration problem. By integrating it with the improved bimodal motion planner, we form a complete TABV exploration framework.



\section{Problem Statement and System Overview}
\label{sec:statement}
\changed{The goal of the proposed method is to explore an initially unknown but bounded 3D space using a TABV under a given energy budget \( E_{\text{all}} \) and time budget \( T_{\text{all}} \). Importantly, the objective is \emph{not} full coverage of the environment, but to collect as much informative data as possible and ensure that the robot can return to the departure station with the data, which is a more reliable approach in practical application such as communication-denied post-disaster environments.
}

\changed{To achieve this, we formulate the exploration task as selecting a sequence of viewpoints and their associated modalities that maximizes the perception of unknown space, while ensuring execution within the available energy and time budgets. Formally, the problem is defined as:
}
\changed{
\begin{subequations}
	\label{eq:main_opt_problem}
	\begin{align}
		\bm{\mathcal{P}}^{*} &= \operatorname{argmax}_{\bm{\mathcal{P}}} IG(\bm{\mathcal{P}}) \\
		\text{s.t.} \quad E_r(\bm{\mathcal{P}}) &\ge 0, \\
		T_r(\bm{\mathcal{P}}) &\ge 0,
	\end{align}
\end{subequations}
}
\changed{where \( \bm{\mathcal{P}} = \{ \mathcal{P}_i \mid i = 0,\dots,n \} \) is a sequence of selected viewpoints. \( IG(\bm{\mathcal{P}}) \) measures the information gain along the trajectory, and \( E_r(\bm{\mathcal{P}}) \), \( T_r(\bm{\mathcal{P}}) \) denote the remaining energy and time after visiting all selected viewpoints and returning to the departure station.}

\changed{
To cope with the inequality constraints and environmental uncertainty, we convert (\ref{eq:main_opt_problem}) into an unconstrained optimization problem using a penalty function approach:
}
\changed{
\begin{equation}
\label{eq:opt_problem}
\bm{\mathcal{P}}^{*} = \operatorname{argmin}_{\bm{\mathcal{P}}} \left( -IG(\bm{\mathcal{P}}) + \kappa_{E_r}(E_r(\bm{\mathcal{P}})) + \kappa_{T_r}(T_r(\bm{\mathcal{P}})) \right),
\end{equation}
}
\changed{
where \( \kappa_{E_r}(\cdot) \) and \( \kappa_{T_r}(\cdot) \) are exponential penalty terms based on remaining energy and time, which will be detailed in Sec.~\ref{subsec:Tree Structure}.
Considering the practical scenario, exceeding the energy limit may prevent return and thus risk mission failure, while moderate time overruns are more tolerable. Therefore, we adopt a steeper penalty curve for energy, while the time-related penalty is relatively moderate and designed to take effect only when the remaining energy is already sufficient. This formulation encourages safe and efficient planning by maintaining a margin to cope with environmental uncertainty.
}


An overview of our TABV autonomous exploration system is presented in Fig.~\ref{fig:cluster}. 
We first extract the exploration information from environment and generate potential viewpoints (Sec.~\ref{sec:Viewpoints Generation}). 
Then we conduct BM-MCTS to determine the traverse sequence of the viewpoints (Sec.~\ref{sec:BM-MCTS}) under the given energy and time budget. Finally, we use the bimodal motion planner to generate bimodal trajectories (Sec.~\ref{sec:Motion Planner}).

\section{Exploration Information Extraction}
\label{sec:Viewpoints Generation}

\subsection{Bimodal Viewpoints Generation}
\label{sec:vp}
Similar to classic frontier-based exploration, frontiers are defined as free voxels adjacent to unknown space. \changed{As the robot moves, new frontiers appear. The robot iteratively plans viewpoints to cover frontiers and replans as the map updates.}

\changed{
To efficiently generate viewpoints covering frontiers, we first group the frontier voxels into clusters.
For each frontier cluster $C$, we consider two strategies for generating viewpoints to achieve full coverage:
i) \textbf{Aerial-Modal Strategy (AS)}: The cluster is covered exclusively using aerial viewpoints.
ii) \textbf{Hybrid-Modal Strategy (HS)}: The cluster is first covered using terrestrial viewpoints; if full coverage cannot be achieved, additional aerial viewpoints are selected to complete the coverage.
The AS and HS sets provide diverse modality-specific candidate viewpoints, which serve as inputs for the subsequent decision-making in BM-MCTS.}

\changed{
Specifically, as shown in Fig.~\ref{fig:cluster}(b), for each $C$, we first sample raw candidate terrestrial and aerial viewpoints. 
Raw aerial viewpoints $\bm{\mathcal{P}}_\text{aerial}$ are generated using cylindrical coordinate sampling around $C$. Raw terrestrial viewpoints $\bm{\mathcal{P}}_\text{ground}$ are sampled from nearby traversable ground voxels extracted from the grid map, with a fixed vertical offset.
Given these raw viewpoints, the problem of selecting bimodal viewpoints to cover $C$ exhibits submodularity \cite{nemhauser1978analysis}. We apply a greedy method to efficiently solve it and obtain AS and HS viewpoint sets. Details and algorithm are provided in the Sec.~1 of the supplementary material \cite{yuman_gao_2025_15510722}.
}

\changed{
Finally, we obtain two sets of candidate viewpoints representing alternative coverage strategies:
$\bm{\mathcal{P}}_\text{AS}$ for pure aerial-modal coverage, and
$\bm{\mathcal{P}}_\text{HS}$ for hybrid-modal coverage.
Each viewpoint is defined as $\mathcal{P}_i = (\mathbf{p}, \phi)$, where $\mathbf{p}$ is the position and $\phi$ is the yaw angle.
\changed{The $IG$ of $\mathcal{P}_i$ is defined as the number of visible frontier voxels within $\mathcal{P}_{i}$'s FoV.}
All viewpoints belonging to the same cluster compose a viewpoint group $\mathcal{G}=\{\bm{\mathcal{P}}_\text{AS}, \bm{\mathcal{P}}_\text{HS} \}$.
}



\begin{algorithm}[t]
	\DontPrintSemicolon
	
	\SetKwInput{KwInput}{Input}                
	\SetKwInput{KwOutput}{Output}              
	\DontPrintSemicolon
	
	\SetKwFunction{FMain}{Main}
	\SetKwFunction{FSum}{Sum}
	\SetKwFunction{FSub}{Sub}
	
	\SetKwProg{Fn}{Function}{:}{}
	\SetKwProg{MFn}{Main Function}{:}{}
	\MFn{{Search($v_0$)}}{
		\While{\changed{number of iterations} is less than threshold}{
			$v_{s}$ = \textit{Select}($v_0$)\;
			[\textit{succ}, $v_{e}$] = \textit{Expand}($v_{s}$, Reward)\;
			\If{succ}
			{
				\textit{Simulate}($v_e$)\;
				\textit{BackPropogate}($v_{s}$, $v_{e}$)
			}
		}
		\KwRet \textit{BestChild}($v_0$)\;
	}
	
	\caption{BM-MCTS}
	\label{al1}
\end{algorithm}

\begin{figure}[t]
        \vspace{-0.5cm}
	\begin{center}
		\includegraphics[width=1.0\columnwidth]{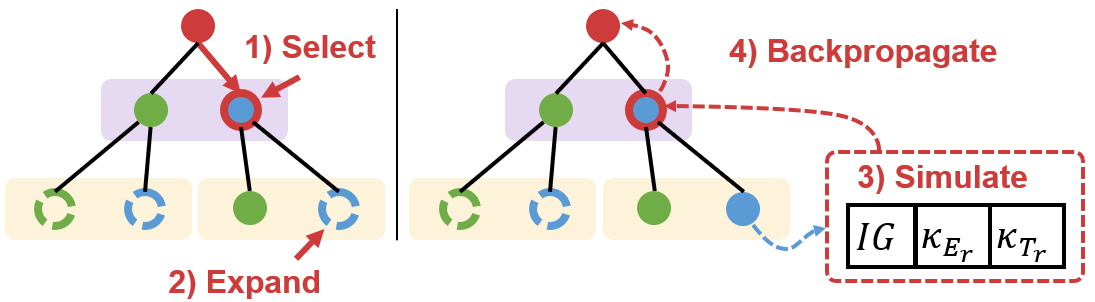}
	\end{center}
	\vspace{-0.0cm}
	\caption{
		\label{fig:vp_to_tree} An illustration of the process of BM-MCTS.
	}
	\vspace{-1.5cm}
\end{figure}

\subsection{Energy and Time Cost between Viewpoints}
\label{sec:cost}
As we choose $E_r$ and $T_r$ as criteria to achieve energy- and time-aware planning, the energy and time cost should be modeled for the estimation process. Given two viewpoints $\mathcal{P}_i$ and $\mathcal{P}_j$, the time cost between them is defined as:
\begin{equation}
	\begin{gathered}
		T(\mathcal{P}_i,\mathcal{P}_j,M)=\max \left\{\frac{\textit{length} (\mathcal{P}_i,\mathcal{P}_j)}{v_{M,\max }}\right. ,
		\left.\frac{\textit{dyaw}(\mathcal{P}_i,\mathcal{P}_j)}{\omega_{\mathrm{M,max}}}\right\},
	\end{gathered}
\end{equation} 
where $\textit{length}(\mathcal{P}_i,\mathcal{P}_j)$ is the length of the searched path between $\mathcal{P}_i$ and $\mathcal{P}_j$, $\textit{dyaw}(\mathcal{P}_i,\mathcal{P}_j)$ is the minimum yaw angle difference between the two viewpoints, $v_{M,\max }$ and $\omega_{\mathrm{M,max}}$ are maximum speed and maximum yaw angular speed in modality $M$, respectively.
For the energy cost, we use a simple constant power model, and the energy cost between them is defined as:
\begin{equation}
	E(\mathcal{P}_i,\mathcal{P}_j,M)=P_{M} \cdot T(\mathcal{P}_i,\mathcal{P}_j,M),
\end{equation}
where $P_{M}$ is the average power under modality $M$. $M$ denotes the modality: $M\in \{ T,A \}$. $T$ and $A$ indicate terrestrial modality and aerial modality, respectively. Specially, we denote $M=(T+A)/2$ as the average modality which means $v_{(T+A)/2,\max }=(v_{T,\max }+v_{A,\max })/2$ and $P_{(T+A)/2}=(P_{T}+P_{A})/2$. 

And if the time and energy cost between $\mathcal{G}$s need to be estimated, taking $E(\mathcal{G}_i,\mathcal{G}_j,M)$ for example, we use an average viewpoint with the mean position and yaw angle of all viewpoints in $\mathcal{G}$ for calculation.

According to existing bimodal vehicle design papers \cite{fan2019autonomous, kalantari2020drivocopter,10538378}, the average power in aerial modality is about $5 \sim 8 \times$ higher \changed{($7.2 \times$ for ours)} than terrestrial modality. 
And the maximum speed in terrestrial modality is about $1m/s$. However, the maximum speed in aerial modality \changed{can} not be achieved $5 \sim 8 \times$ faster in exploration due to safety and perception accuracy. 
So the terrestrial modality has an advantage in energy cost, while the aerial modality results in less time cost.


\section{Bimodal Monte Carlo Tree Search}
\label{sec:BM-MCTS}


The Monte Carlo Tree Search \cite{coulom2006efficient, Liu-RSS-19} is a planning method for finding the optimal decision within a given horizon under limited computational resources. 
Based on that, we propose an extended MCTS method, called BM-MCTS, to select the optimal viewpoint sequence with two potential modalities. 
The BM-MCTS method consists of four key steps: selection, expansion, simulation, and backpropagation, as shown in Fig.~\ref{fig:vp_to_tree}.
After the required iterations, the best child node of the root is selected as the next local goal.
And the best branch corresponds to the \changed{resulting} bimodal path.
The whole algorithm is shown as Alg.~\ref{al1}.

\begin{figure}[t]
	\begin{center}
		\includegraphics[width=0.9\columnwidth]{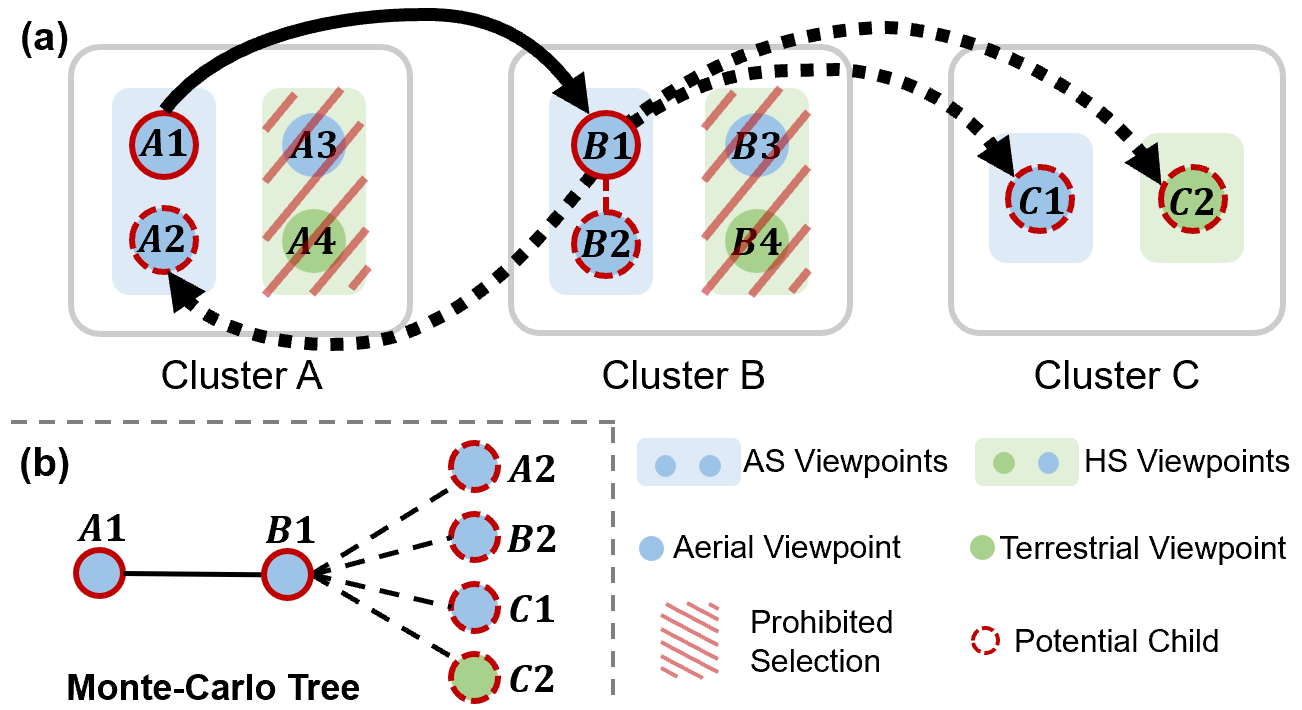}
	\end{center}
	\vspace{-0.0cm}
	\caption{
		\label{fig:poten_child} An example for determining the potential children. (a): The potential children of the viewpoint $B1$ of the frontier cluster $B$. If the viewpoint belongs to a cluster that has already been expanded, then the viewpoints with the other modality are prohibited from selection. (b): The corresponding Monte Carlo tree. Each branch represents a viewpoint traversal sequence.
	}
	\vspace{-1.5cm}
\end{figure}

\subsection{Tree Structure and Reward}
\label{subsec:Tree Structure}

First, we define the structure of the tree. 
Each tree node \changed{$v$} corresponds to a viewpoint \changed{$\mathcal{P}$}, and each branch determines a viewpoint traversal sequence, as shown in Fig.~\ref{fig:poten_child}(b).

\changed{Each node \( v \) is associated with several attributes. Specifically, \( E_R(v) \) and \( T_R(v) \) denote the estimated remaining energy/time when the robot reaches the viewpoint \( \mathcal{P} \) corresponding to node \( v \), by following the sequence of viewpoints along the current search branch.  
\( E_r(v) \) and \( T_r(v) \) represent the estimated remaining energy/time \emph{after} the robot reaches \( \mathcal{P} \), completes visiting the remaining selected viewpoints, and returns to the departure station.}

\begin{figure}[t]
	\begin{center}
        \vspace{-0.0cm}		\includegraphics[width=0.9\columnwidth]{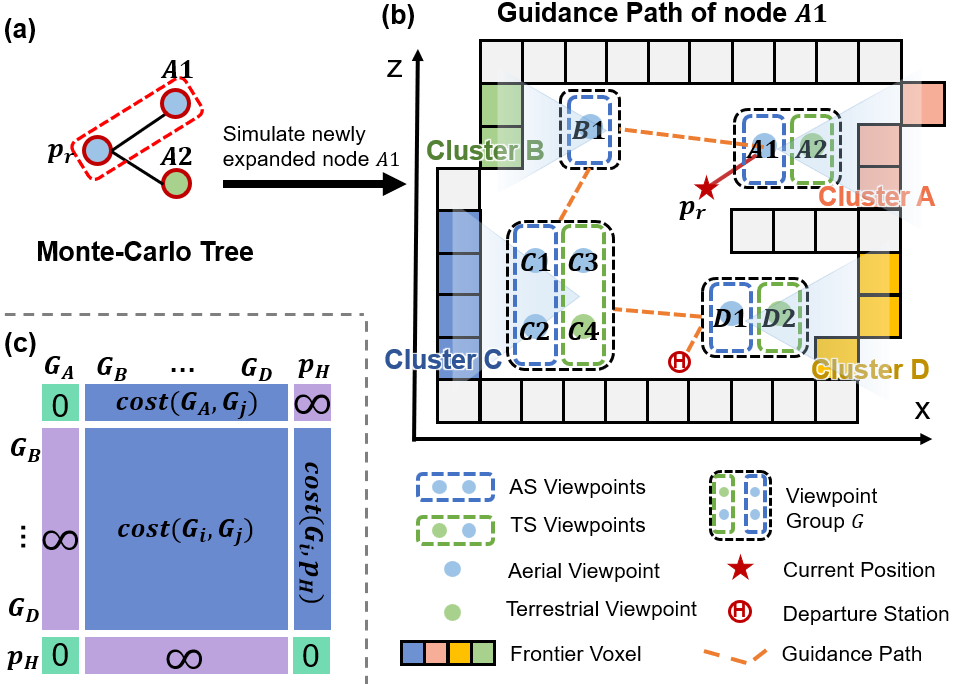}
	\end{center}
	\vspace{-0.2cm}
	\caption{
		\label{fig:tsp} 
		An example of the \changed{guidance} path generation for a newly expanded node.
		(a): The expanding Monte Carlo tree, where node $A1$ is newly expanded and requires simulation. 
		(b): The \changed{guidance} path generation for node $A1$. In this case, $A1$ is chosen in $\mathcal{G}_A$ to cover cluster $A$, and the path from robot's position $p_r$ to $A1$ is determined. Then we solve a grouped TSP to get the whole \changed{guidance} path that traverses all clusters and returns home.
		(c): The cost matrix of grouped TSP. The purple region denotes the infinity connection, and the green region denotes the zero connection.
	}
	\vspace{-1.0cm}
\end{figure}

As for connections between nodes, it is necessary to specify the potential children of each node.
In the case of a viewpoint $\mathcal{P}_i$, its potential child $\mathcal{P}_j$ should satisfy the following conditions:
1). $\mathcal{P}_j$ is not already included in the branch of $\mathcal{P}_i$. 
2). The distance between $\mathcal{P}_i$ and $\mathcal{P}_j$ is smaller than the threshold. 3). If the viewpoint group to which $\mathcal{P}_j$ belongs has been expanded in the branch of $\mathcal{P}_i$, $\mathcal{P}_j$ should maintain the same sampling mode, as shown in Fig.~\ref{fig:poten_child}.
\changed{If no candidate $\mathcal{P}_j$ satisfies all the conditions, we relax the second condition to allow feasible selection.}

Second, we define the reward of the nodes for tree search.
As formulated in \changed{(\ref{eq:opt_problem})}, the reward function consists of \emph{process gain}, represented by $IG$, and \emph{terminal cost}, represented by $E_r$ and $T_r$.
The process gain is related to all nodes in the subtree of this node. While the terminal cost is only related to the leaf nodes in the subtree of this node.
Each node $v$ has reward $\mathbf{R}(v) = [R_p(v), R_t(v)]$, where $R_p$ is the process gain and $R_t$ is the terminal cost:
\begin{subequations}
	\label{eq:reward}
	\begin{align}
		&R_p(v) = IG(v)/{n_\text{\changed{IG}}}(v),\\
		&R_t(v) = \kappa_{E_r}(E_r(v)) + \kappa_{T_r}(T_r(v)),
	\end{align}
\end{subequations}
\changed{where $n_\text{IG}(v)$ denotes the discounted visitation count of node $v$, which is incremented by a fixed weight $\gamma_{\text{IG}}=0.8$ during each backpropagation step} (Alg.~\ref{alg:Backward}). 
\changed{Correspondingly, $IG(v)$ accumulates the discounted $IG$ from its child nodes during backpropagation.}
\changed{
$R_p(v) = IG(v) / n_\text{IG}(v)$ therefore reflects the discounted average $IG$ across the subtree rooted at $v$. This formulation mitigates the bias caused by uneven subtree sizes and emphasizes
information from nearby nodes.}
$\kappa_{E_r}(E_r(v))$ and $\kappa_{T_r}(T_r(v))$ denote the average terminal energy and time cost computed over all \emph{leaf} nodes in the subtree rooted at $v$.
The costs are constructed using the exponential functions defined as
$\kappa_{E_r}(x)=exp(-a_1 \cdot x/E_{all}+b_1)$, $\kappa_{T_r}(x)=exp(-a_2 \cdot x/T_{all}+b_2)$, where $a_i,b_i$ are preset hyperparameters.


	
	
	
	

	
	

\begin{figure*}[t]
	\begin{center}
		\includegraphics[width=2.0\columnwidth]{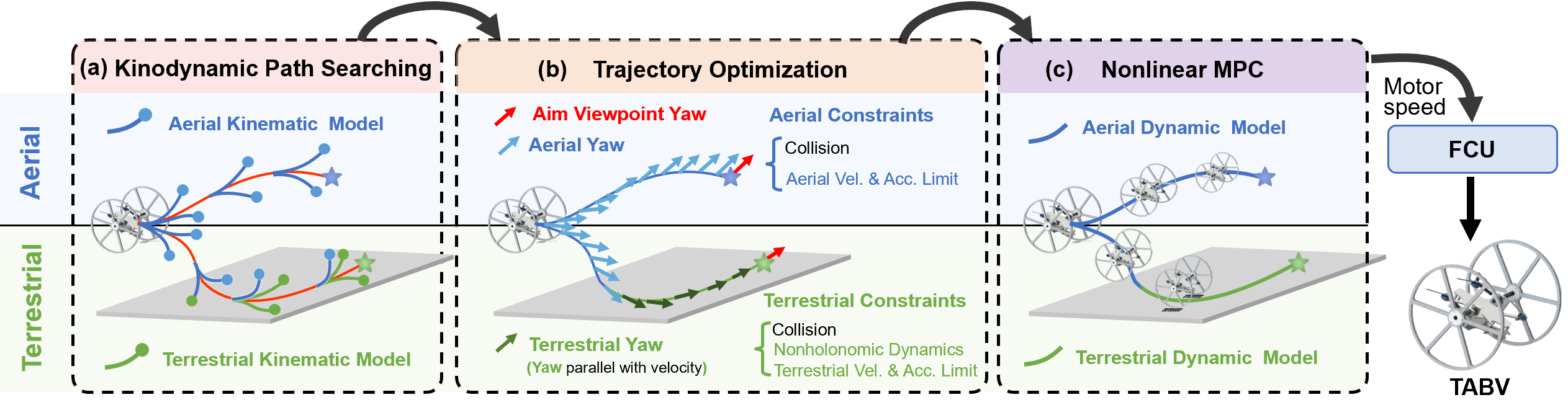}
	\end{center}
	\vspace{-0.2cm}
	\caption{
		\label{fig:planning} The hierarchical bimodal motion planning framework. (a): The kinodynamic path search front end. (b): The flatness-differential-based spatial-temporal trajectory optimization back end. (c): The NMPC module \cite{zhang2023model} to calculate the desired motor speed.
	}
	\vspace{-0.6cm}
\end{figure*}

\subsection{Selection and Expansion Process}
In the selection process, we recursively choose the optimal node from the root until reaching a node with unexpanded potential children. The selection policy follows the upper confidence bound (UCB) rule \cite{Liu-RSS-19}, balancing exploration and exploitation. In the function \textit{BestChild}($v$), we compute the UCB score $\boldsymbol{U}(v_k)$ for each child node $v_k$ of $v$:
\begin{subequations}
	\label{eq:traj opt problem}
	\begin{align}
		&\boldsymbol{U}(v_k)=G(v_k)\changed{-}\sqrt{2 \ln n_s/n(v_{k})},\\
		&G(v_k) = \changed{-}N(R_p(v_k)) + R_t(v_k), 
	\end{align}
\end{subequations}
where \changed{$n(v_k)$ is the number of times $v_k$ has been selected so far}. 
$n_s=\sum_{k=1}^{K}n(v_{k})$, \changed{where $K$ is the number of $v$'s child nodes}.
$N(x)$ \changed{linearly maps} $x$ to $[\epsilon,1]$, where $\epsilon = 0.05$ represents the smallest normalized gain. 
Among all child nodes, the one with the minimum $\boldsymbol{U}(\cdot)$ is selected.

Given the selected node $v_{i}$ to be expanded corresponding to viewpoint $\mathcal{P}_{i}$, its potential children are determined as Sec.~\ref{subsec:Tree Structure}. A child $v_{i+1}$ is randomly selected from $v_{i}$'s unexpanded potential children.
Then we update the left energy $E_R$ and left time $T_R$ when robot reaches $v_{i+1}$:
\begin{equation}
	\begin{aligned}
		E_R(v_{i+1}) = &E_R(v_{i}) 
		- E(\mathcal{P}_{i}, \mathcal{P}_{i+1},M(\mathcal{P}_{i+1})),
	\end{aligned}
\end{equation}
where $E(\mathcal{P}_{i}, \mathcal{P}_{i+1},M(\mathcal{P}_{i+1}))$ is the energy consumption from $\mathcal{P}_{i}$ to $\mathcal{P}_{i+1}$, and the modality is determined by the latter one.
As $E_r$ and $T_r$ are handled similarly, we only present the energy-related equations here and in the following sections.

\subsection{Simulation Process with \changed{Guidance} Path}
The simulation process updates the reward of the newly expanded node, which requires evaluating its $IG$, $E_r$, and $T_r$.
For $IG$, $IG(v_{i})$ is initialized as the number of \changed{visible} frontier voxels within $\mathcal{P}_{i}$'s FoV, serving as an estimate of newly gathered information.
To estimate the energy and time required for traversing through all clusters and returning home, we solve an extended grouped Traveling Salesman Problem to generate a guiding path, as shown in Fig.~\ref{fig:tsp}.

We design the cost matrix, as shown in Fig.~\ref{fig:tsp}(c), to set the departure station $p_H$ as the final destination.
The cost between viewpoint groups is defined as the travel time between them, given by $T(\mathcal{G}_i, \mathcal{G}_j, (T+A)/2)$. Since the exact modality of nodes is unknown during evaluation, we set $M = (T+A)/2$.

\begin{figure*}[t]
	\vspace{0.0cm}
	\centering
	\begin{center}
		\includegraphics[width=2.0\columnwidth]{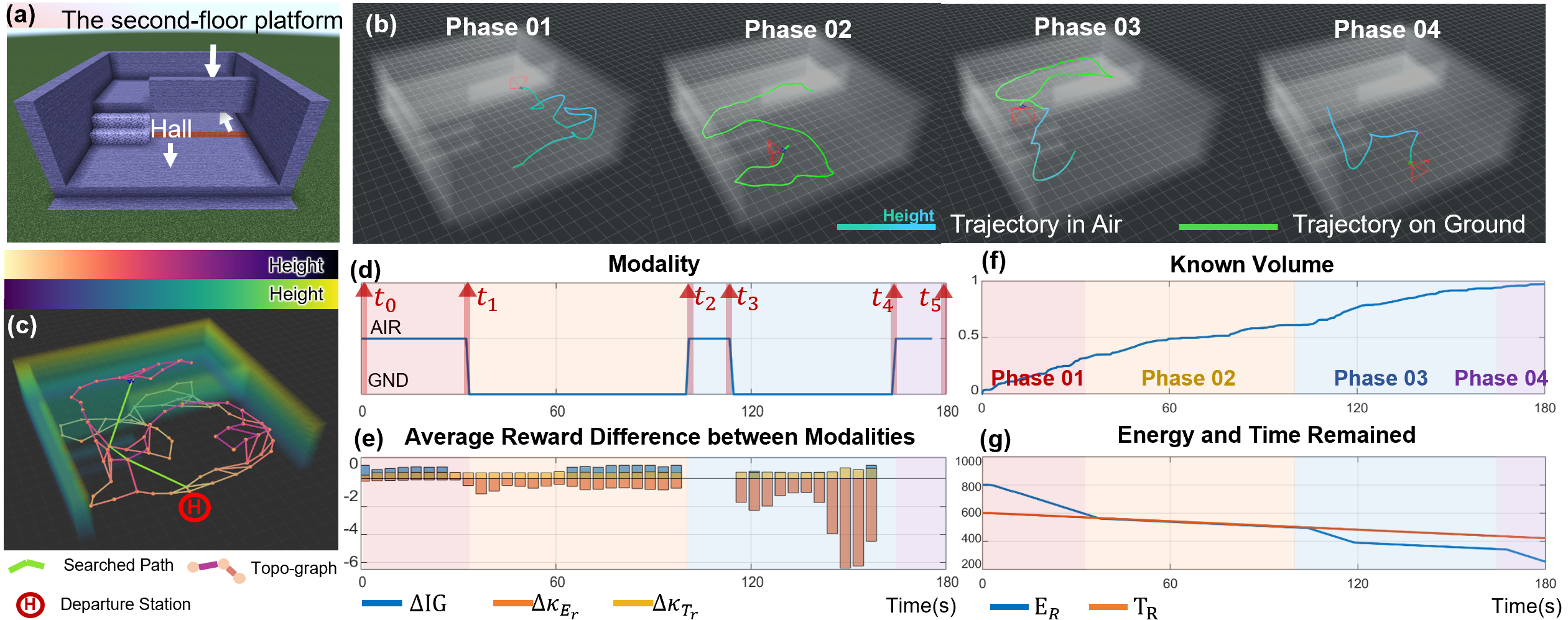}
	\end{center}
	\vspace{-0.0cm}
    \caption{
    (a) A two-story house scene for TABV exploration.  
    (b) Four exploration phases. 
    \textit{Phase 01}: Fly to explore most of the hall.  
    \textit{Phase 02}: Cover the first-floor platform, then fly upstairs.  
    \textit{Phase 03}: Roll to cover the second-floor platform.  
    \textit{Phase 04}: Cover the remaining hall and return home.  
    (c) Topo-graph illustration.  
    (d) TABV modalities.  
    (e) Average reward difference between aerial and terrestrial child nodes at the second depth in the Monte Carlo tree.  
    (f) Coverage ratio over time.  
    (g) Remaining energy and time.}
	\label{fig:2layer}
	\vspace{-0.6cm}
\end{figure*}

The main computational bottleneck in the simulation process is the cost matrix calculation.
To efficiently estimate the feasible path length, we maintain a global topo-graph that records the visited positions at intervals and connects nodes within a distance threshold, as \changed{shown in Fig.~\ref{fig:2layer}(c)}. By performing an $A^{\star}$ search on the topo-graph, we obtain a fast conservative path length estimation.
Moreover, paths between viewpoint groups are incrementally updated during viewpoint generation, further improving efficiency.

Then the simulation step is executed to update $E_r$, $T_r$:
\begin{equation}
	\begin{aligned}
		E_r(v_{i}) =& E_R(v_{i}) -
		\sum_{q=i}^{k-1}E(\mathcal{G}_{q},\mathcal{G}_{q+1},(T+A)/2)\\
        &- E(\mathcal{G}_{k},\mathbf{p}_{H}),
	\end{aligned}
\end{equation}
where $v_{i}$ corresponds to $\mathcal{P}_i \in \mathcal{G}_{i}$, and the last two terms of the equation represent the estimated energy consumption for the robot traveling from the current viewpoint group to the last one and returning home according to the \changed{guidance} path.

\subsection{Backpropagation Process}

The obtained reward should be backpropagated to each visited node to update the reward of those nodes, preparing for the next iteration of selection.
As the process gain $R_p$ is related to all the nodes on the same branch, the cumulative backpropagation is adopted. While the terminal cost $R_t$ is only related to the leaf node on the branch, we adopt the average backpropagation for it.
The backpropagation process is detailed in Alg.~\ref{alg:Backward}.


\begin{algorithm}[b]
	\DontPrintSemicolon
	
	\SetKwInput{KwInput}{Input}                
	\SetKwInput{KwOutput}{Output}              
	\DontPrintSemicolon
	
	\SetKwFunction{FMain}{Main}
	\SetKwFunction{FSum}{Sum}
	\SetKwFunction{FSub}{Sub}
	
	\SetKwProg{Fn}{Function}{:}{}
	\SetKwProg{MFn}{Main Function}{:}{}
	
	\Fn{{BackPropagate($v$, $v_k$)}}{
        $n(v)$ = $n(v)$ + $1$\;
		\changed{$n_\text{IG}(v)$ = $n_\text{IG}(v)$ + $\gamma_{\text{IG}}$ \;}
		$IG(v)$ = $IG(v)$ + $\gamma_{\text{\changed{IG}}} IG(v_k)$\;

		$\kappa_{E_r}(v)$ = $\sum_{v_c\in v.kid} \kappa_{E_r}(v_c)$ / Num($v.kid$) \;
		$\kappa_{T_r}(v)$ = $\sum_{v_c\in v.kid} \kappa_{T_r}(v_c)$ / Num($v.kid$) \;
		
		\textit{BackPropogate}($v.parent$, $v$)
	}
	\caption{Backward Process}
	\label{alg:Backward}
\end{algorithm}

\subsection{Prune Condition}
To reduce the search space and improve efficiency, \changed{a child node is pruned and no longer expanded if the remaining energy after reaching node $v_i$ is less than the energy required to return home from $v_i$:}
\changed{
\begin{equation}
    E_R(v_{i}) - E(\mathcal{G}_{i},\mathbf{p}_{H})<0,
\end{equation}
}
\changed{where $v_{i}$ corresponds to $\mathcal{P}_i \in \mathcal{G}_{i}$.}

\changed{Further details of algorithm analysis are presented in Sec.~5 of the supplementary material \cite{yuman_gao_2025_15510722}.}

\section{Bimodal Motion Planning}
\label{sec:Motion Planner}

Given the next goal from BM-MCTS, we use a hierarchical bimodal motion planning method to generate terrestrial-aerial hybrid trajectories and control the TABV to execute.
Building upon our previous work \cite{zhang2023model}, we briefly summarize the overall pipeline here and highlight the key enhancements.

As shown in Fig.~\ref{fig:planning}, the planner follows a standard hierarchical architecture with a kinodynamic path search front end and a spatial-temporal trajectory optimization back end. To better support our exploration framework, we introduce several key enhancements.

\changed{First, instead of assuming a fixed ground plane, we perform online terrain perception using incremental ground segmentation. This allows the system to dynamically identify traversable surfaces in complex, multi-level environments without relying on predefined structural assumptions.}

\changed{Second, we introduce modality-aware planning to replace the prior approach that consistently favored the terrestrial modality.} In the front end, motion primitives are selected according to the target modality: aerial primitives for aerial targets, and bimodal primitives with penalties on aerial motion for terrestrial targets. In the back end, modality-specific constraints are applied—such as nonholonomic dynamics for terrestrial segments. Additionally, we also integrate bimodal yaw planning into the back end.  

Lastly, to enhance safety, we compute the Euclidean Signed Distance Field (ESDF) for each ground segment to query the distance to edges, ensuring the TABV flies safely near edges and prevents falls.


Finally, by incorporating this bimodal motion planner, a complete autonomous terrestrial-aerial exploration framework is established.
\changed{More details are provided in the Sec.~2 of the supplementary material \cite{yuman_gao_2025_15510722}.}

\section{Result}


\subsection{Simulation}
To validate our TABV exploration system, we conduct simulations and phased analyses in multiple multilayered buildings. Moreover, we analyze the adaptability of the BM-MCTS under different budgets, as well as the relationship between solution quality and iteration times. 
Based on our TABV platform, the power of the aerial modality is 7 times that of the terrestrial modality.
So we set $P_T=1$, and $P_A = 7$, \changed{meaning that one second of movement in terrestrial mode consumes 1 unit of energy, while one second in aerial mode consumes 7 units. To avoid redundancy, we omit the units for time and energy in the remainder of this section.}
Moreover, the maximum velocity is set as $v_{T,\max}=0.5~\mathrm{m/s}$, $v_{A,\max}=1.0~\mathrm{m/s}$.




\begin{figure*}[h]
    \centering
    \includegraphics[width=2.0\columnwidth]{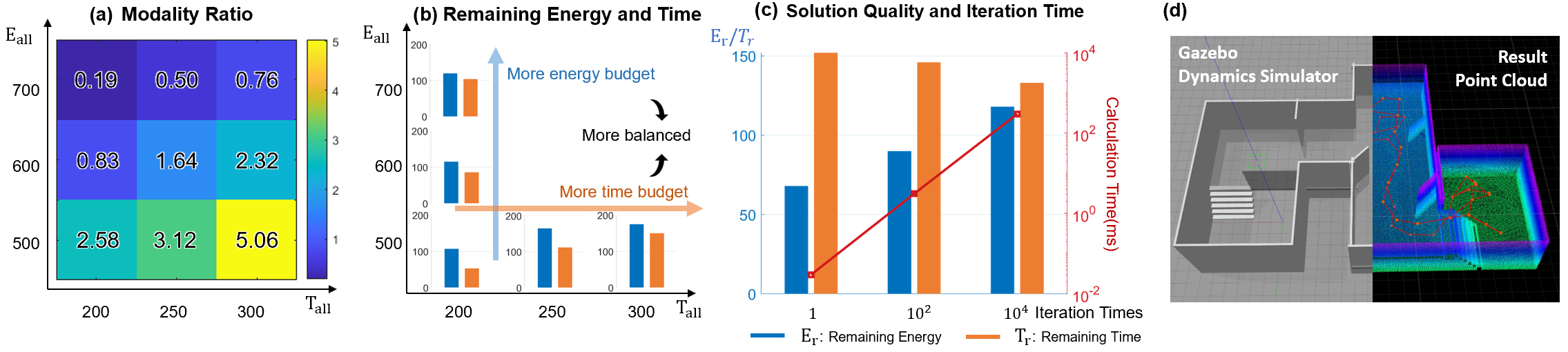}
    \caption{(a)-(b): Performance under different energy/time budgets. (a): The ratio of modality under different budgets, where the value represents the ratio of time on the ground to time in the air. (b): The remaining energy/time when exploration is completed. (c): The algorithm calculation time and the remaining energy/time when exploration is completed under different iteration times. \changed{(d): More simulation scenario and exploration result provided in the supplementary material \cite{yuman_gao_2025_15510722}.}}
    \label{fig:map2_data}

    \vspace{0.3cm}  

    \includegraphics[width=2.0\columnwidth]{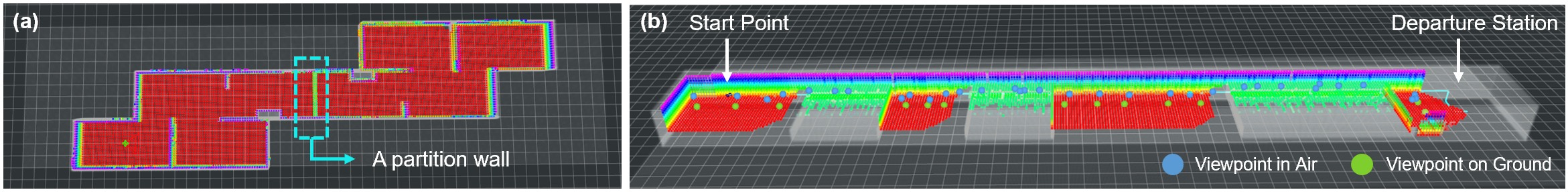}
    \caption{Simulation scenes. (a): A multi-room scene with a 0.5m height partition wall. (b): A scene with a series of viewpoints.}
    \vspace{-0.5cm}
    \label{fig:map2}
\end{figure*}

\subsubsection{\textbf{Bimodal Exploration for A Two-story House} }
We construct a $15~\mathrm{m}\times 15~\mathrm{m} \times 6~\mathrm{m}$ two-story house in MineCraft as an exploration scene (Fig.~\ref{fig:2layer}(a)). The first floor of the house is $2m$ high, connected by a staircase to a second-floor platform. 
The energy budget is set to 1000, and the time budget is set to 600. The hyper-parameters $\kappa_{E_r}$ and $\kappa_{T_r}$ \changed{are set identically to those described in Sec.~9 of the supplementary material \cite{yuman_gao_2025_15510722}.} In Fig.~\ref{fig:2layer}(e), the average reward difference is defined as $\Delta \bar{\mathbf{R}}=\bar{\mathbf{R}}(v_A)-\bar{\mathbf{R}}(v_G)$, where $\bar{\mathbf{R}}(v_A)$ (or $\bar{\mathbf{R}}(v_G)$) is the average reward of the root node's aerial (or terrestrial) modality child nodes.

We divided the whole exploration process into four phases for detailed analysis (Fig.~\ref{fig:2layer}). In Phase 01, both of the remaining energy and time are sufficient, so the $IG$ reward takes advantage. The TABV switches to aerial modality for more information gain along with a shorter time. At $t_1$, the TABV gets into the first-floor platform. As the platform height is only $2$ m, the $IG$ reward is almost the same under two modalities. Since a lot of energy is consumed in Phase 01, energy counts more than time $t_1$, leading the TABV to choose the terrestrial modality for the entire first-floor platform. At $t_2$, after covering all terrestrial viewpoints, the TABV takes off and lands on the second floor at $t_3$, then rolls to cover the second-floor platform to save energy. In Phase 04, with only aerial viewpoints remaining in the hall, the TABV flies to cover them and returns home.


\subsubsection{\textbf{Adaptability to Different Budgets}}
We show the performance of our method under different energy and time budgets. The simulation scene is a $40~\mathrm{m}\times 16~\mathrm{m} \times 2~\mathrm{m}$ office-like area, which is completely separated by a $1~\mathrm{m}$ high partition wall (as shown in Fig.~\ref{fig:map2}(a)). 
We define the modality ratio as the time on the ground divided by the time in the air.
Experiments are conducted under varying energy and time budgets, with five trials for each setting. The results are shown in Fig.~\ref{fig:map2_data}. 
As the time budget \changed{increases} and the energy budget \changed{decreases}, TABV tends to favor the terrestrial modality. We further analyze the remaining energy and time at the end of exploration, as shown in Fig.~\ref{fig:map2_data}(b).
With the same energy budget, a larger time budget results in more remaining energy; with the same time budget, a larger energy budget results in more remaining time. 
This demonstrates that the increasing one provides greater flexibility for more efficient use of the other, highlighting the flexibility of the BM-MCTS method in balancing energy and time, and enabling the TABV to adapt to different budgets.

\subsubsection{\textbf{Solution Quality and Iteration Times}}
Since the BM-MCTS method can produce solutions at any time when reaching the iteration threshold, we analyze the relationship between the iteration times and the quality of the solution.
We built a one-way scene as Fig.~\ref{fig:map2}(b) is shown to ensure consistent exploration direction in the test.
The TABV first flies along a preset trajectory from the departure station to the start point to generate a series of bimodal viewpoints. 
Then the exploration starts with energy and time budget set to 400 and 200, respectively. For each preset iteration time, we simulate for ten times. 
As shown in Fig.~\ref{fig:map2_data}(c), as the maximum iteration time increases, the remaining energy and time become more balanced.
This is because the energy and time consumption estimates become more accurate as the search tree expands, at the cost of increased calculation time.

\changed{Further simulation tests and comprehensive comparisons are presented in Sec.~6-8 of the supplementary material \cite{yuman_gao_2025_15510722}.}

\begin{figure*}[t]
	\vspace{-0.0cm}
	\centering
	\begin{center}
		\includegraphics[width=2.0\columnwidth]{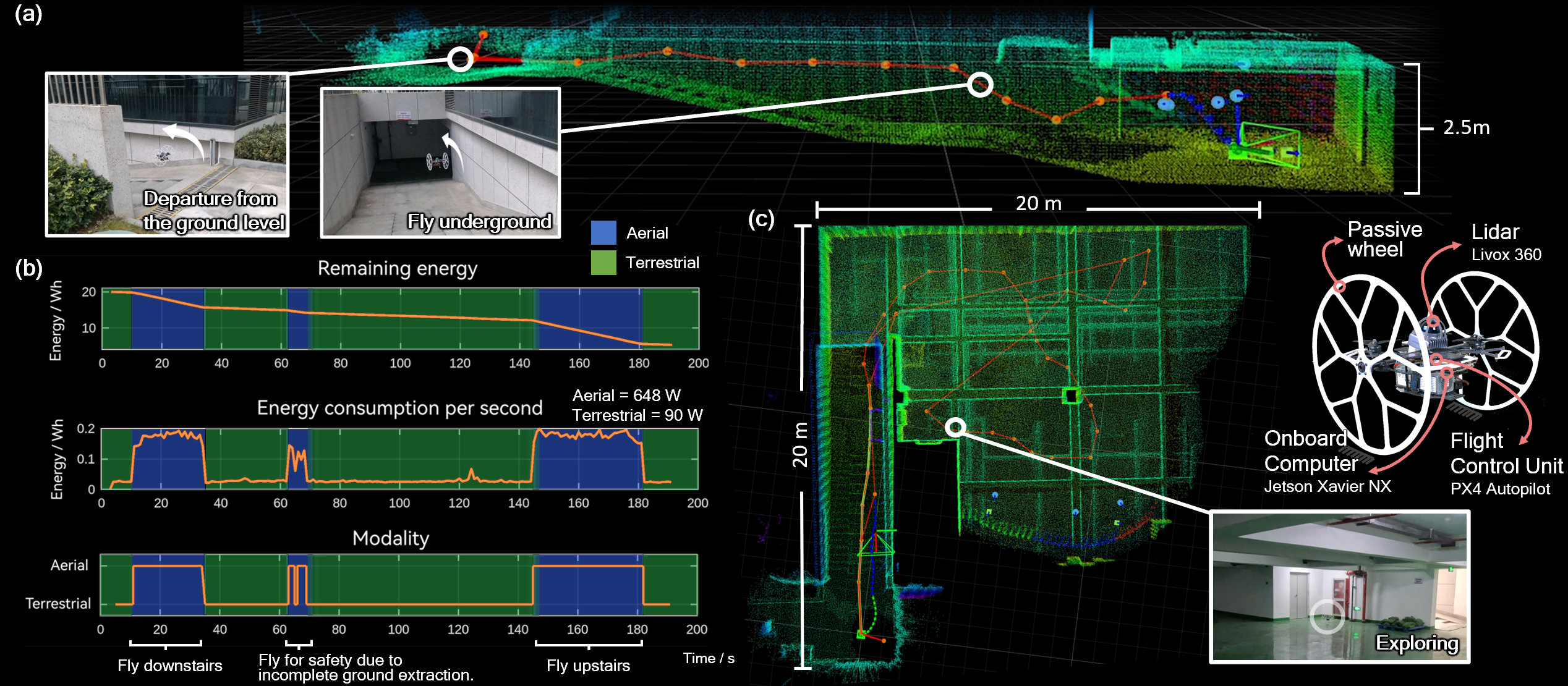}
	\end{center}
	\vspace{-0.0cm}
	\caption{Real-world experiment. (a): Side view of the entrance to the underground parking garage. (b): The energy and modality curves. (c): Top view of the explored map. }
	\label{fig:real}
	\vspace{-0.0cm}
\end{figure*}
\subsection{Real-World Experiment}
To demonstrate and verify the proposed approaches in
real-world environments, we use a customized TABV platform,
as shown in Fig.~\ref{fig:real}. The TABV weighs $1.85$~kg, equipped with a Livox 360 and a Jetson Xavier NX. We use FAST-LIO2 \cite{xu2022fast} for real-time localization. All the algorithms are running onboard.
The TABV consumes $648$ W in the aerial modality and $90$ W in the terrestrial modality, making the aerial power consumption $7.2$ times higher.
We set a $20$~Wh serve energy budget and a $900$ s abundant time budget for the TABV to explore an underground parking garage. 
The TABV presents the adaptation to fly across uneven stairs and switch to terrestrial modality to explore with low energy cost.
Finally, the TABV returns home safely when the energy budget is used up.



\section{Conclusion}
\label{sec:conclusion}
In this paper, we develop a hierarchical scheme to drive the TABV to explore under given energy and time budget. With this scheme, the TABV can flexibly respond to different environments and energy/time constraints by changing the modality. \changed{A detailed analysis of the system’s limitations is provided in Sec.~11 of the supplementary material\cite{yuman_gao_2025_15510722}.} For future work, we will add environment prediction for more accurate energy/time consumption estimation.

\bibliography{RAL2025}
\clearpage


\section*{Supplementary material}

Supplementary Materials for
# Autonomous Exploration with Terrestrial-Aerial Bimodal Vehicles

Yuman Gao,[*] Ruibin Zhang,[*] Tiancheng Lai, Yanjun Cao, Chao Xu, Fei Gao[†]

## 1 Bimodal Viewpoint Generation

In this section, we present the bimodal viewpoint generation pipeline and describe a greedy algorithm that exploits submodularity to efficiently select informative viewpoints.

To efficiently generate viewpoints covering frontiers, we first group the frontier voxels into clusters. Frontier cells are clustered into frontier clusters based on their 3D spatial distribution. For a frontier cluster $C$, we consider two strategies for selecting viewpoints to achieve full coverage:

i) **Aerial-Modal Strategy (AS)**: The cluster is covered exclusively using aerial viewpoints.

ii) **Hybrid-Modal Strategy (HS)**: The cluster is first covered using terrestrial viewpoints; if full coverage cannot be achieved, additional aerial viewpoints are selected to complete the coverage.

The AS and HS sets provide diverse modality-specific candidate viewpoints, which serve as informative inputs for subsequent decision-making in BM-MCTS. The detailed procedure is presented in Alg.1 and illustrated in Fig.1.

Specifically, for each frontier cluster $C$, we first sample raw candidate terrestrial and aerial viewpoints (Lines 1–2). As shown in Fig.1(a), raw aerial viewpoints are generated using cylindrical coordinate sampling within a distance range $[d_{min}, d_{max}]$ and azimuth angle $\varphi$ around the cluster's normal, across multiple heights along the $z$-axis. The frontier normal computed via Principal Component Analysis (PCA) and oriented toward the known space. Constraining the sampling range in this manner improves efficiency without sacrificing coverage quality. While, raw terrestrial viewpoints are generated by selecting ground voxels from the nearby grid map and applying a fixed vertical offset. To ensure spatial diversity, a minimum distance is enforced between adjacent terrestrial viewpoints. Importantly, each

---
[*]Equal contribution. All authors are with the Institute of Cyber-Systems and Control, College of Control Science and Engineering, Zhejiang University, Hangzhou 310027, China, and also with the Huzhou Institute, Zhejiang University, Huzhou 313000, China.

[†]Corresponding author: Fei Gao. Emails: {ymgao,fgaoaa}@zju.edu.cn

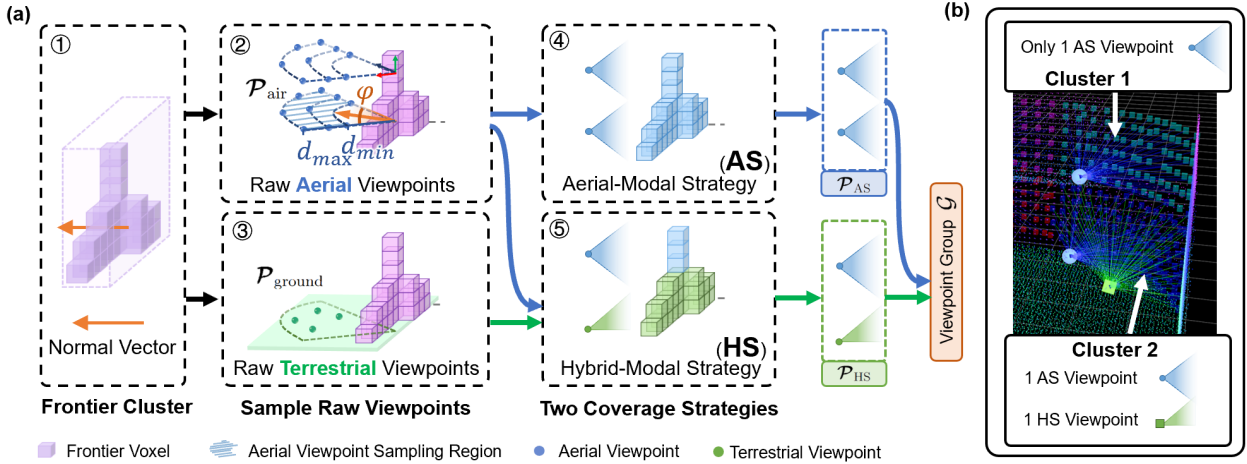

Figure 1: (a): The pipeline of bimodal viewpoint generation with two coverage strategies. ①: A frontier cluster $C$ with a normal computed via PCA and oriented toward the known space. ②-③: Raw aerial viewpoints $\boldsymbol{\mathcal{P}}_{\mathrm{aerial}}$ and raw terrestrail viewpoints $\boldsymbol{\mathcal{P}}_{\mathrm{ground}}$ generation. ④-⑤: Two strategies to select bimodal viewpoints to fully cover $C$, resulting $\boldsymbol{\mathcal{P}}_{\mathrm{AS}}$ and $\boldsymbol{\mathcal{P}}_{\mathrm{HS}}$. (b). An example of AS and HS viewpoints generation in the simulation.

viewpoint is defined as $\mathcal{P}_i = (\mathbf{p}, \phi)$, where $\mathbf{p}$ is the position and $\phi$ is the yaw angle. The information gain (IG) of $\mathcal{P}_i$ is defined as the number of visible frontier voxels within $\mathcal{P}_i$'s Feild of View (FoV). The yaw angle $\phi$ is determined as the one maximizing IG, by using a yaw optimization method similar to [1].

Given these raw viewpoints, we then select the viewpoint to fully cover $C$ via two strategies: AS and HS. For AS, which covers the cluster using only aerial viewpoints, in each iteration, the viewpoint with the highest newly gained information $\Delta IG$ is selected, where $\Delta IG$ is defined as the number of previously unseen frontier voxels observable by the candidate viewpoint. The iteration continues until either the overall coverage of the cluster exceeds 95%, or the maximum $\Delta IG$ falls below 15% of the total frontier cell number of $C$ (Line 3-11). The selection of viewpoints under HS follows the same procedure, except that terrestrial viewpoints are prioritized. If the selected terrestrial viewpoints alone cannot achieve sufficient coverage, aerial viewpoints are subsequently introduced to complete the coverage (Line 12-27).

This process produces two alternative viewpoint sets $\boldsymbol{\mathcal{P}}_{\mathrm{AS}}$ and $\boldsymbol{\mathcal{P}}_{\mathrm{HS}}$ each corresponding to a valid coverage strategy. All viewpoints belonging to the same cluster compose a viewpoint group $\mathcal{G} = \{\boldsymbol{\mathcal{P}}_{\mathrm{AS}}, \boldsymbol{\mathcal{P}}_{\mathrm{HS}}\}$. An example of the viewpoints generated under AS and HS is shown in Fig. 1(b).

Note that the problem exhibits submodularity [2]; that is, as more viewpoints are selected, the marginal IG from selecting an additional viewpoint decreases. Nemhauser et al. [2] proved that a greedy algorithm—which starts from the empty set and iteratively selects the element with the highest marginal gain—guarantees a solution with total reward no less than $(1 - 1/e)$ of the optimal value. Therefore, we adopt a greedy strategy for viewpoint selection in both AS and HS procedures.

---
**Algorithm 1** Generate Bimodal Viewpoints for Cluster $C$
---
**Input:** Frontier cluster $C$
**Output:** AS viewpoints, HS viewpoints

1  // **Raw Viewpoints Sampling**
2  $\mathcal{P}_{\text{ground}} \leftarrow$ `sampleRawGroundViewpoints`$(C)$
3  $\mathcal{P}_{\text{air}} \leftarrow$ `sampleRawAirViewpoints`$(C)$

4  // **Aerial-Modal Strategy (AS)**
5  $\mathcal{P}_{\text{AS}} \leftarrow \emptyset,\ coverage \leftarrow 0$
6  **while** $coverage < 0.95$ **do**
7      $\mathcal{P} \leftarrow$ viewpoint in $\mathcal{P}_{\text{air}}$ with max $\Delta$IG
8      **if** $max\ \Delta IG < 0.15 \times C.size$ **then**
9          **Break**
10     $\mathcal{P}_{\text{AS}} \leftarrow \mathcal{P}_{\text{AS}} \cup \{\mathcal{P}\}$
11     update coverage with $\mathcal{P}$
12     $\mathcal{P}_{\text{air}} \leftarrow \mathcal{P}_{\text{air}} \setminus \{\mathcal{P}\}$

13 // **Hybrid-Modal Strategy (HS)**
14 $\mathcal{P}_{\text{HS}} \leftarrow \emptyset,\ coverage \leftarrow 0$
15 **while** $coverage < 0.95$ **do**
16     $\mathcal{P} \leftarrow$ viewpoint in $\mathcal{P}_{\text{ground}}$ with max $\Delta$IG
17     **if** $max\ \Delta IG < 0.15 \times C.size$ **then**
18         **Break**
19     $\mathcal{P}_{\text{HS}} \leftarrow \mathcal{P}_{\text{HS}} \cup \{\mathcal{P}\}$
20     update coverage with $\mathcal{P}$
21     $\mathcal{P}_{\text{ground}} \leftarrow \mathcal{P}_{\text{ground}} \setminus \{\mathcal{P}\}$
22 **while** $coverage < 0.95$ **do**
23     $\mathcal{P} \leftarrow$ viewpoint in $\mathcal{P}_{\text{air}}$ with max $\Delta$IG
24     **if** $max\ \Delta IG < 0.15 \times C.size$ **then**
25         **Break**
26     $\mathcal{P}_{\text{HS}} \leftarrow \mathcal{P}_{\text{HS}} \cup \{\mathcal{P}\}$
27     update coverage with $\mathcal{P}$
28     $\mathcal{P}_{\text{air}} \leftarrow \mathcal{P}_{\text{air}} \setminus \{\mathcal{P}\}$
29 **return** $\mathcal{P}_{\text{AS}}$, $\mathcal{P}_{\text{HS}}$
---

# 2   Enhancements of Bimodal Motion Planner

To better support our exploration framework, we introduce several key enhancements over our previous design.

First, instead of assuming a predefined ground plane (e.g., at a fixed height $z = 0$), we employ an online ground extraction module based on 3D perception, which dynamically

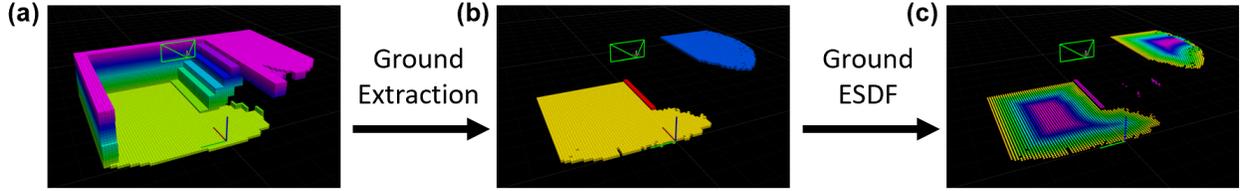

Figure 2: Ground extraction and ground ESDF generation. (a). Grid map. (b). Extracted ground part from the grid map. (c). ESDFs of the ground segments.

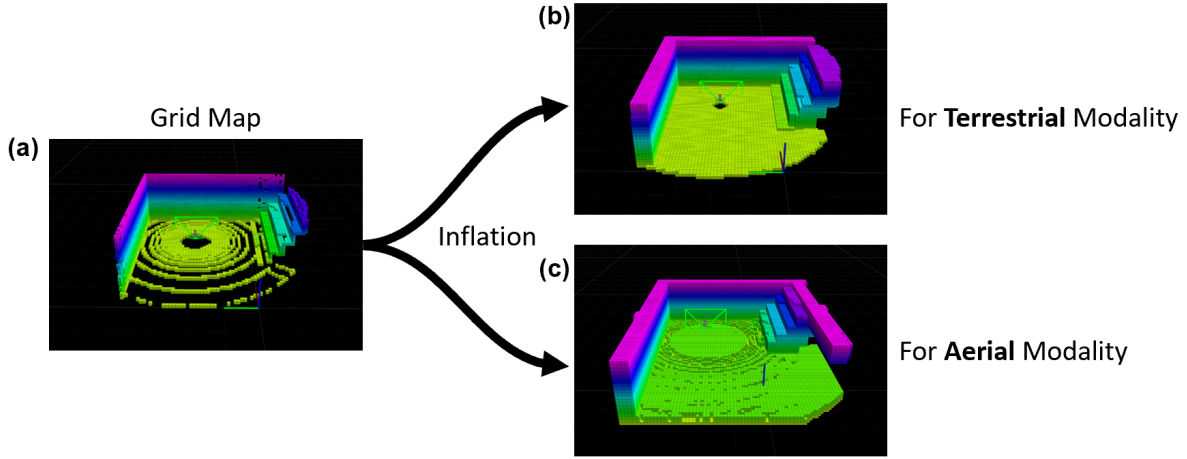

Figure 3: Grid map generation for different modalities. (a). The voxel grid map before inflation. (b). The inflated grid map for terrestrial path searching. The z axis of ground segments will not be inflated. (c). The inflated grid map for aerial path searching.

segments traversable surfaces from the environment, as shown in Fig. 2(b). Specifically, we extract ground segments from the voxel grid map using an incremental breadth-first search (BFS). A voxel is identified as a ground cell if it has a sufficient number of neighboring voxels at a similar height within a local window. A newly detected ground cell will inherit the ground segment ID from the neighboring old one. Fig. 2(b) shows two ground segments with two different IDs. When a new portion of the grid map is built, we use BFS to continue expanding ground segments from neighboring ground-consistent cells. Importantly, we check each ground segment, and if the number of connected voxels is too small, it will not be treated as valid ground—such as the small red patch in Fig. 2(b). The online terrain perception enables the planner to generate feasible paths in complex, multi-level environments without relying on prior structural assumptions.

Second, rather than discarding all point cloud data below a certain height threshold to avoid constructing a ground-influenced grid map—as done in our previous work—we now build two separate voxel grid maps tailored for each modality as shown in Fig. 3. For the terrestrial modality, the ground part of the grid map will not inflated along the $z$ axis, enabling the terrestrial path searching to search closely on the ground. For aerial modality, we will perform standard isotropic inflation in all three dimensions, enabling collision-free aerial path searching.

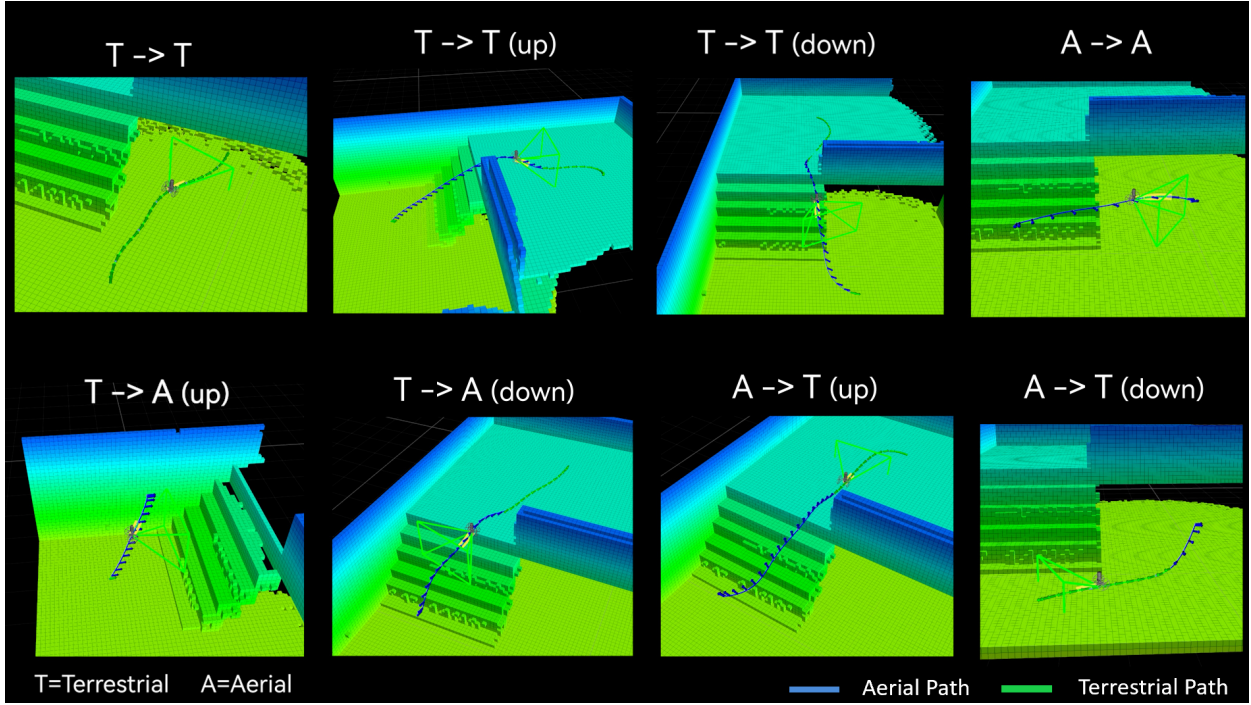

Figure 4: Results of the bimodal motion planner in simulation.

Third, unlike the previous method that always preferred the terrestrial modality by applying a constant penalty to aerial actions, our approach handles the two modalities separately. In the kinodynamic path search front end, we expand motion primitives based on the modality of the target: for aerial target, only aerial primitives are used; for terrestrial target, both terrestrial and aerial primitives are expanded, with penalties on aerial ones to enable switching only when necessary (e.g., stair climbing or obstacle overcrossing). In the back end, we apply modality-specific constraints: aerial segments are bounded by aerial velocity and acceleration limits, while terrestrial segments additionally obey nonholonomic dynamics. The back-end design is the same as our prior work [3].

Lastly, to improve safety near discontinuities in terrain, we compute the Euclidean Signed Distance Field (ESDF) for each ground segment as shown in Fig. 2(c). The ESDF provides distance-to-edge information for back-end trajectory optimization, allowing the system to plan safe aerial trajectories near boundaries and avoid unintended falls. As shown in the terrestrial-to-terrestrial (T→T (down)) trajectory in Fig. 4, the TABV switches to the aerial modality before the edge to ensure safety.

Altogether, these enhancements allow the TABV to perform motion planning in unknown, multi-level environments, resulting in a fully functional and autonomous bimodal exploration system. We present the improved bimodal motion planning in simulation (Fig. 4) and real world (Fig. 5).

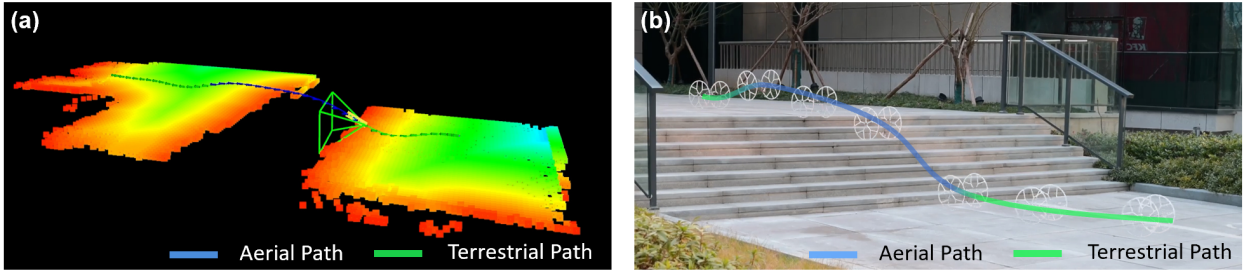

Figure 5: Real-world multi-layer motion planning experiment. (a). The planned bimodal trajecotry and ESDF of extracted ground. (b). The snapshot of the real-world experiment.

# 3 Topological Graph

## 3.1 Construction of the Topological Graph

We construct and maintain a topological graph (topo-graph) during exploration to record the spatial connectivity traversed by the robot. This allows efficient global path queries, even between distant frontier clusters, by estimating navigation distances directly on the graph.

The topo-graph is built using a simple rule: a new node is added every fixed distance along the robot's trajectory. For each new node, we identify nearby nodes within a predefined Euclidean distance threshold. If a nearby node is visible from the current node (determined via ray-casting in the grid map) and the topological distance between them on the existing graph exceeds a threshold, an edge is created between the two. This process is then recursively applied to the neighbors of the connected node, enabling sparse yet meaningful graph connectivity and avoiding overly dense links.

Additionally, each new node is always connected to the most recent node in the topo-graph, ensuring continuity along the robot's path.

An result of topo-graph is presented in Fig. 6.

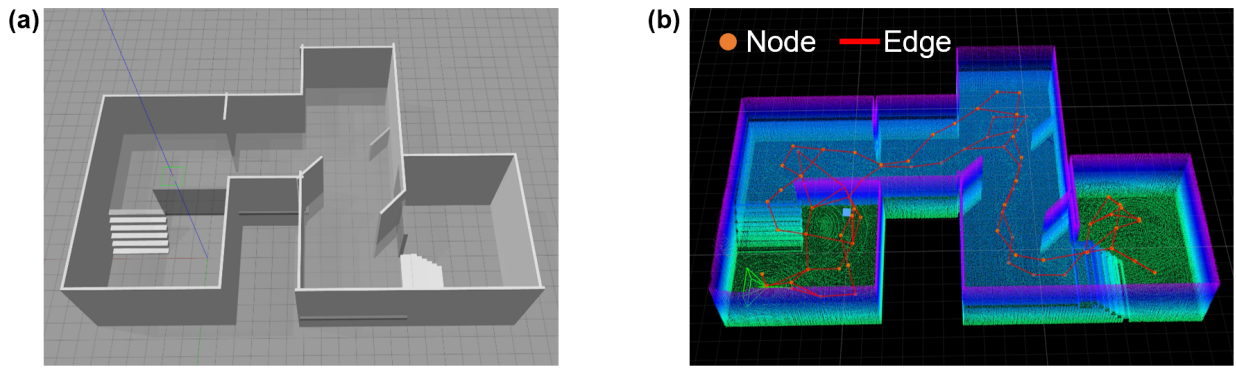

Figure 6: (a). The exploration scene in the Gazebo simulator. (b). The result of topo-graph at the end of exploration.

## 3.2 Path Searching using Topological Graph

By performing an $A^\star$ search on the topo-graph, we obtain a fast conservative path length estimation. Specifically, when estimating the path length between two viewpoints (In the simulation step in BM-MCTS), we use a three-stage strategy based on their Euclidean distance and search time constraints:

i) If the straight-line distance is less than a predefined threshold (`dis_thres`), we first attempt an $A^*$ search on the voxel grid map for a more accurate estimate. If this search exceeds a time threshold without success, we fall back to topological graph search.

ii) If the straight-line distance exceeds the threshold, or if the voxel map search fails due to timeout, we perform a faster $A^*$ search on the topological graph. This provides a coarser but efficient approximation of the path.

iii) If both methods fail to find a feasible path, we conservatively assign a large cost value to indicate infeasibility.

This hybrid estimation strategy balances accuracy and computational efficiency, supporting real-time performance in the BM-MCTS framework.

# 4 Replanning and Exit Mechanism

The estimated energy and time consumption may not fully reflect real-world factors such as terrain resistance or aerodynamic effects, leading to inaccuracies. Additionally, as the robot moves, new information becomes available. Therefore, it is essential to *replan* in real time. Replanning is triggered under the following conditions:

**i)** The selected local goal lies in a frontier cluster whose coverage exceeds 50%, meaning the majority of its frontier cells no longer qualify as frontier cells (i.e., they are no longer free cells adjacent to unknown space).

**ii)** The previously planned trajectory has been fully executed.

**iii)** A predefined time interval has elapsed.

In addition, the system must also decide when to terminate exploration and return to the departure station. This decision is triggered under the following circumstances:

**i)** *No valid solution is found by BM-MCTS.* This occurs when all nodes are pruned, indicating that there is no cluster that can be visited while still preserving sufficient energy and time to return the departure station.

**ii)** *The remaining energy or time is insufficient to safely return.* Specifically, the remaining energy is no greater than $E(\mathcal{G}_i, \mathbf{p}_H)$ or the remaining time is no greater than $T(\mathcal{G}_i, \mathbf{p}_H)$, where both are conservative estimates of the cost to return to the departure station. We adopt a *conservative return strategy* that always uses the same modality as recorded in the topological graph to ensure the feasibility and safety of the return.

iii) *No frontier clusters remain* in the environment, indicating that all unknown regions have been explored.

# 5 BM-MCTS Algorithm Analysis

Note that in the simulation process, we assume $M = (T + A)/2$ to generate a coarse guidance path, introducing inaccuracies. However, the estimation inaccuracy is progressively corrected during the planning process. If the estimation $E_r$ is less than the true value, the tree will not tend to expand through it. But due to the exploration property of the UCB rule, each node has a nonzero probability of being selected and further estimated. If the estimation $E_r$ is overestimated, the node is more likely to be selected to expand. After expansion, the overestimation will be reduced. And the converege of such selection policy is proved [4].

As for the time complexity of the path searching algorithm, the path between $\mathcal{G}$s needs to be computed only when new frontier clusters are generated. The path between $\mathcal{P}$s is computed during the tree expansion only for the newly generated $\mathcal{P}$s. The time complexity of both steps is $\mathcal{O}(UV)$, where $U$ represents the number of new $\mathcal{P}$s or $\mathcal{G}$s, and $V$ represents the number of existing ones. Furthermore, a topo-graph is utilized to accelerate this process.

Additionally, the BM-MCTS is an anytime algorithm, meaning that it can produce solutions at any time during its execution and continually improve the quality of these solutions as it is given more computational resources, until the tree is fully expanded.

# 6 Extreme Resource Limitation Case

In this section, we evaluate the performance of our exploration system under an extreme resource-constrained scenario. As illustrated in Fig. 7, the exploration begins with an energy budget of 25 and a time budget of 30, which is significantly lower than the amount required to cover all available viewpoints.

Despite the limited resources, the TABV successfully explores a portion of the environment and returns to the departure station with remaining energy and time of $E_r = 5$ and $T_r = 10$, respectively. The remaining energy is not exactly zero because the remaining frontier clusters are considered unreachable. The estimated energy was insufficient to cover any of these clusters and return to the departure station safely.

# 7 Details of A Large Two-story House Exploration

In this section, we provide additional details about the experiment, which were omitted in the main paper due to space limit. We test our system in the Gazebo with dynamics simulation as Fig. 8 shown. The test environment is a two-story scene of size 18 m × 24 m × 4 m, featuring two staircases. The TABV is equipped with a 360° LiDAR sensor with a vertical FoV of −30° to 30° and a maximum range of 4.5 meters. Note that our method is compatible with any sensor that has a limited FoV. In this simulation, we use a LiDAR with a constrained vertical FoV, while in the simulation study in Sec. VII.A (Fig. 7 of the main

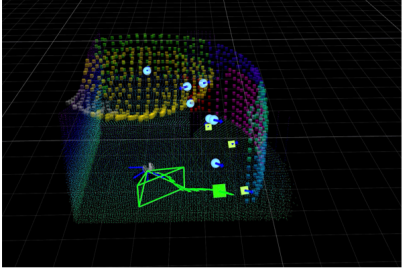 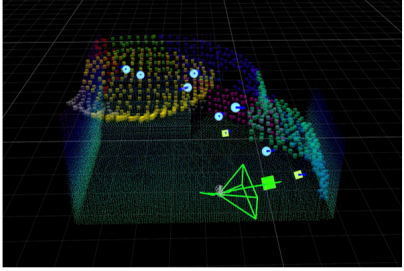 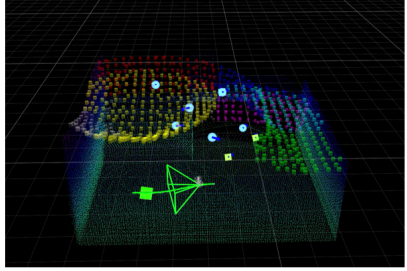

$E_r$ = **22**, T$_r$ = 27    $E_r$ = **17**, T$_r$ = 22    $E_r$ = **9**, T$_r$ = 14

Figure 7: Exploration with energy budget = 25, which much less than the amount for covering all existing viewpoints.

paper), a depth sensor with a horizontal FoV of 90°, vertical FoV of 60°, and a sensing range of 3.5 meters is used. The FoV is explicitly passed as a parameter to the planner, and the viewpoint generation process dynamically adapts to the sensor's FoV.

The exploration task is with an energy budget of 900 and a time budget of 900. During the early stage (from $t_0$ to $t_2$), both energy and time budgets are sufficient, so information gain is prioritized.

- $t_0$–$t_1$: On the first floor, due to its low ceiling, the observation quality of terrestrial and aerial modalities is similar. Since aerial modality consumes more energy, the terrestrial modality is preferred.

- $t_1$–$t_2$: An informative aerial-only viewpoint appears, prompting a switch to the aerial modality. Upon entering a two-story-high hall, aerial viewpoints provide higher information gain, so the robot continues exploring in aerial modality and lands on the second floor after finishing the hall exploration.

- $t_2$–$t_3$: With energy consumption accumulating and remaining budget limited, and aerial modality no longer providing an advantage for information gain, the robot switches back to terrestrial modality to finish exploring the second floor and then flies downstairs.

- $t_3$–$t_4$: A few remaining aerial-only unknown regions near the starting area are covered using the aerial modality.

The entire exploration process is completed in 243 seconds.

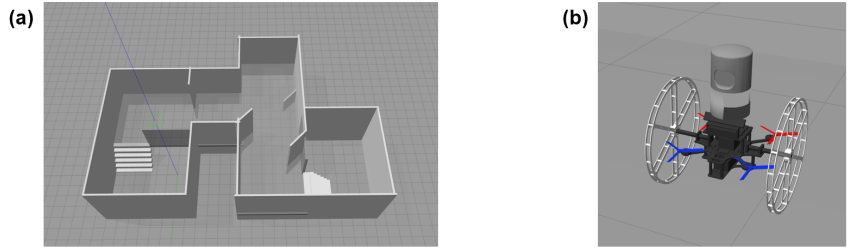

Figure 8: Gazebo simulation. (a). A large two-story house scene. (b). The TABV model with dynamics in Gazebo.

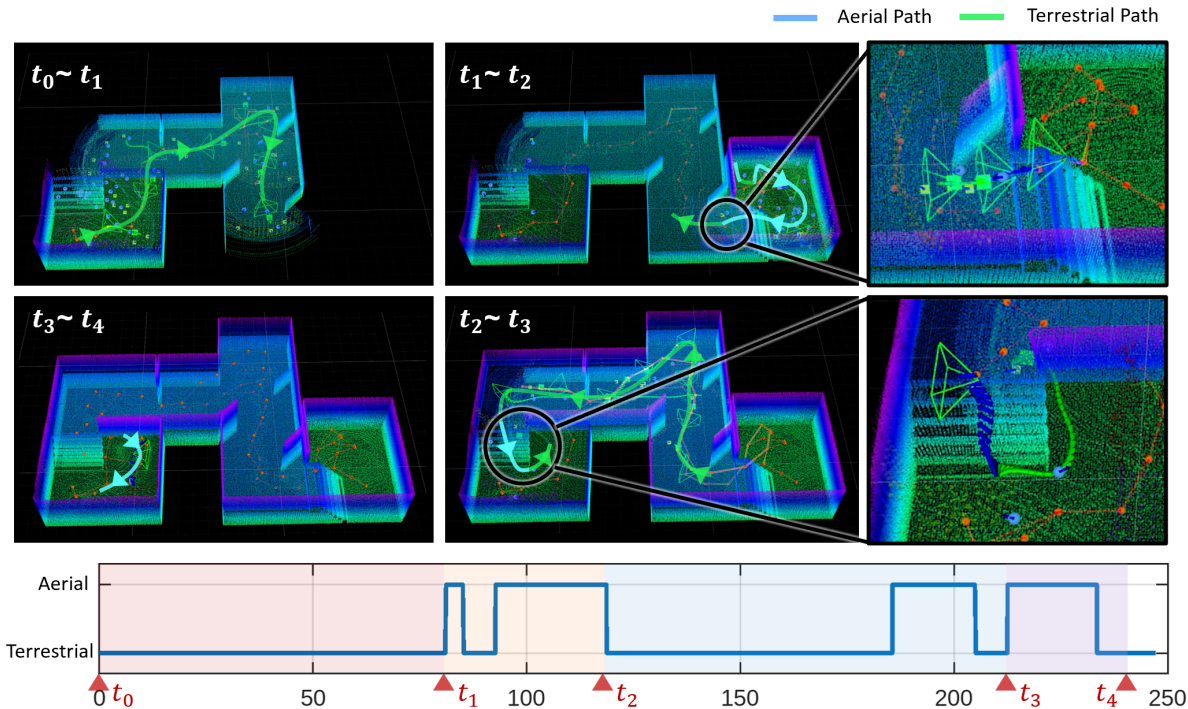

Figure 9: Results of the two-story house exploration.

# 8 Supplementary Comparisons

To better demonstrate the effectiveness of our bimodal exploration system, we present comprehensive comparisons across multiple levels of abstraction. We provide comparisons at three levels:

i) **Framework-level comparison (Sec. 8.1):** Since there is no existing exploration framework specifically designed for bimodal systems, we compare our complete TABV exploration framework against a state-of-the-art purely aerial exploration framework (FUEL [5]). This comparison highlights the advantages of integrating both aerial and terrestrial modalities at the framework level, including improved exploration capability and computational efficiency in large-scale environments.

ii) **Modality configuration comparison (Sec. 8.2):** We conduct ablation studies comparing our bimodal system against three baselines: **UAV-only**, **UGV-only**, and **UGV-cross** (a rule-based bimodal configuration that switches modalities solely based on traversability). This comparison highlights the benefit of integrating both mobility modalities, and making modality-switching based on energy and time constraints.

iii) **Resource-aware decision-making algorithm comparison (Sec. 8.3):** We compare our decision-making algorithm, which optimizes viewpoint sequences under energy and time constraints, against classical methods based on the Travelling Salesman Problem (TSP) and Next-Best-View (NBV) strategies [6]. This highlights the efficiency and adaptability of our planning algorithm in energy and time constrained exploration.

These three levels of comparison jointly validate the contributions of our system design: the integrated framework, the coordinated use of dual modalities, and the planning algorithm under resource constraints.

## 8.1 Exploration Framework-Level Comparison

We perform a framework-level comparison between our bimodal exploration system and FUEL [5], a powerful framework for fast UAV exploration that represents the state-of-the-art in aerial exploration.

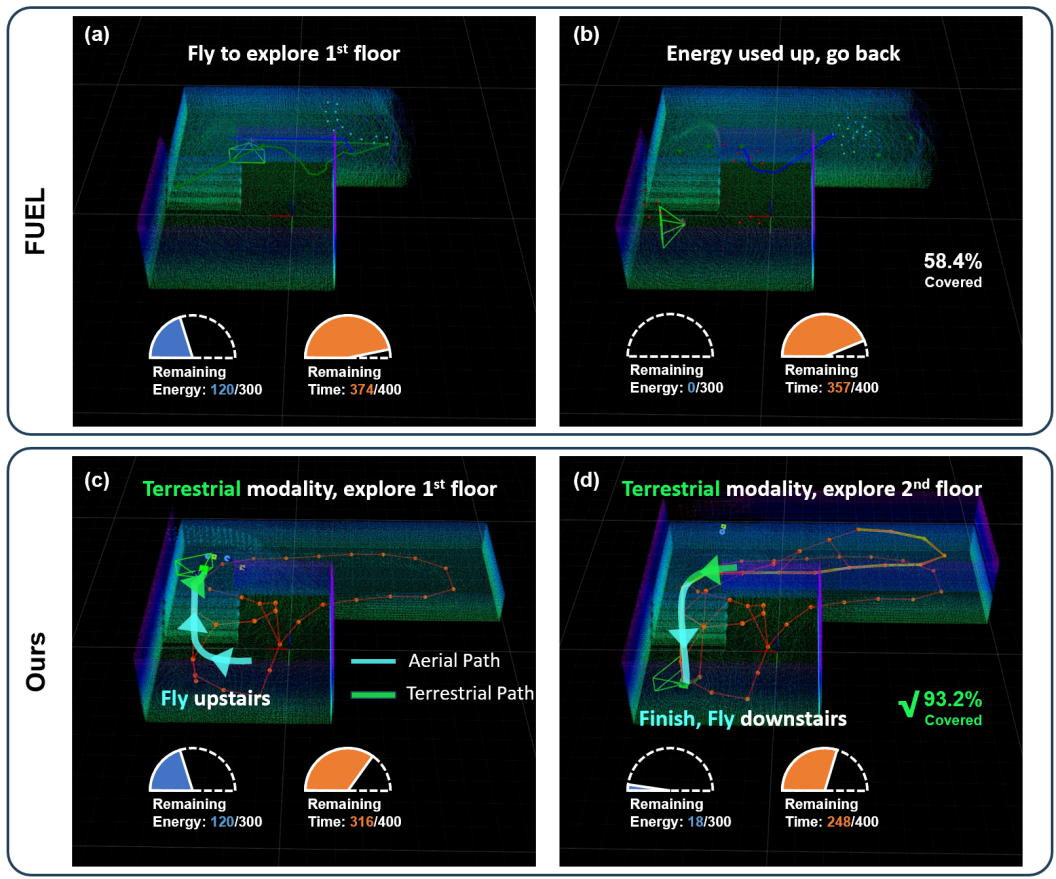

Figure 10: Exploration process of Case 1 (E300, T400) in the simulation. (a)(c). Result when remaining energy = 120. (b)(d). Result when energy is used up.

FUEL is a hierarchical framework designed for purely aerial exploration in complex unknown environments. It incrementally maintains a global Frontier Information Structure (FIS) to support efficient exploration planning. FUEL employs a TSP-based method to generate a global coverage path, followed by local viewpoint refinement and time-optimal trajectory generation. For fair benchmarking, we ported FUEL into our Gazebo simulation and executed its planned trajectories via the same NMPC controller used for TABV. Except for adjusting velocity and acceleration limits as well as the frontier cluster size to match our own system parameters, we did not modify any other components of FUEL. Notably,

Table 1: Framework-Level Comparison

| Case | Method | Coverage (%) | Energy Used | Time Taken (s) | Remarks |
|------|--------|--------------|-------------|----------------|---------|
| Case 1 (E300, T400) | FUEL | 58.4 | 300∗ | 48 | Energy shortage |
| | **Ours** | **93.2✓** | 282∗ | 152 | Almost covered |
| Case 2 (E800, T100) | **FUEL** | **95.4✓** | 700 | 100∗ | Almost covered |
| | **Ours** | 94.6✓ | 460 | 94∗ | Almost covered |

1. ✓ denotes that environment almost covered, ∗ denotes using up the resource.
2. **Bold** entries denote the best results for each metric in the corresponding case.

FUEL does not support ground segmentation, terrestrial viewpoint generation, or decision-making mechanisms for modality switching. Therefore, this section focuses on comparing the overall frameworks. A more detailed comparison of algorithmic performance under the terrestrial-aerial exploration context will be provided in Section 8.3.

As shown in Tab. 1, we compare our exploration framework with FUEL in two cases. In case 1 (E300, T400), under limited energy conditions, FUEL lacks terrestrial mobility and thus fails to complete the exploration task efficiently due to the high energy cost of sustained flight. The exploration details of case 1 are shown in Fig. 10. In case 2 (E800, 100), under tight time constraints, both systems rely primarily on the aerial modality, resulting in comparable performance.

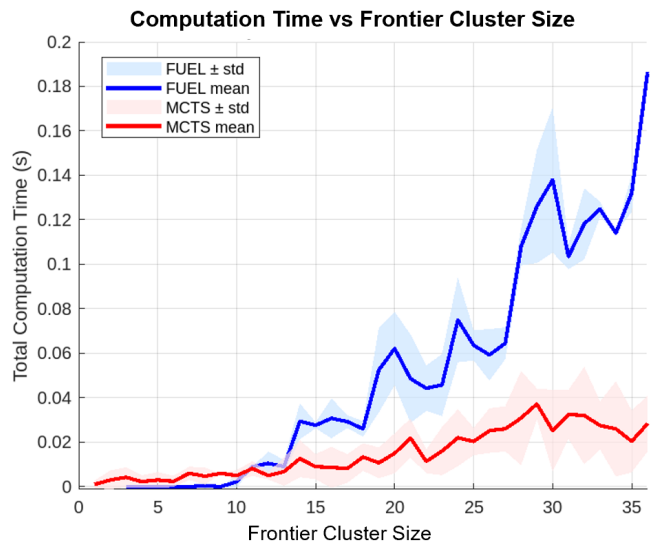

Figure 11: Computation time comparison between FUEL and ours.

Additionally, in large-scale environments, FUEL suffers from low computational efficiency due to the absence of the topo-graph, which is explicitly constructed and utilized in our framework to support scalable planning. The relationship between computation time and the number of frontier clusters is illustrated in Fig. 11. For FUEL, the computation time includes both cost matrix construction and TSP solving; for our method, it accounts for the entire process of Bimodal Monte Carlo Tree Search (BM-MCTS) with ten iterations. The same BM-MCTS parameters are used in the algorithm-level comparison in Sec. 8.3.

In summary, the results shows that the proposed method surpasses FUEL in energy-constraint case and presents comparable performance in time-constraint case. In addition, the proposed MCTS-based approach reveals computational stability against increasing frontier cluster size, while FUEL fails.

## 8.2 Modality Configuration Comparison

To systematically assess the impact of different modality configurations on exploration performance under resource constraints, we conduct ablation studies comparing our proposed bimodal system with three baseline variants:

- **UAV-only**: A single-modality system relying solely on aerial exploration.

- **UGV-only**: A ground-only exploration system, with no aerial capability.

- **UGV-cross**: A hybrid-modal strategy that switches to aerial modality *only* when no viewpoint can be reached via traversable terrestrial paths.

- **Ours**: The full version of our BM-MCTS-based bimodal exploration system with strategic modality switching.

These systems are evaluated in the simulation conditions in two budget-constrained cases. The results are shown in Tab. 2 and Fig. 12. The modification for each variant is only the exploration planner part, and they use the same motion planner.

This comparison also directly responds to the reviewer's concern that our results might be achievable by simply combining the two modalities by flying over untraversable parts of the environment and using terrestrial modality in the rest of the environment (correspnding to the **UGV-cross** variant).

Table 2: Modality Configuration Comparison

| Case | Method | Coverage (%) | Energy Used | Time Taken (s) | Remarks |
|------|--------|--------------|-------------|----------------|---------|
| Case 1 (E300, T400) | UAV-only | 65.1 | 300* | 43 | Energy shortage |
| | UGV-only | 61.5 | 90 | 90 | $2^{nd}$ floor unreachable |
| | **UGV-cross** | **92.1**✓ | **236*** | **67** | Almost covered |
| | **Ours** | **93.2**✓ | **282*** | **48** | Almost covered |
| Case 2 (E800, T100) | **UAV-only** | **94.6**✓ | **460** | **94*** | Almost covered |
| | UGV-only | 61.5 | 96 | 96* | $2^{nd}$ floor unreachable |
| | UGV-cross | 66.4 | 136 | 100* | Time shortage |
| | **Ours** | **94.6**✓ | **460** | **94*** | Almost covered |

1. ✓ denotes that environment almost covered, * denotes using up the resource.
2. **Bold** entries denote the best results for each metric in the corresponding case.
3. UGV-cross: a baseline strategy that switches to aerial modality only when no viewpoint can be reached via traversable terrestrial paths.

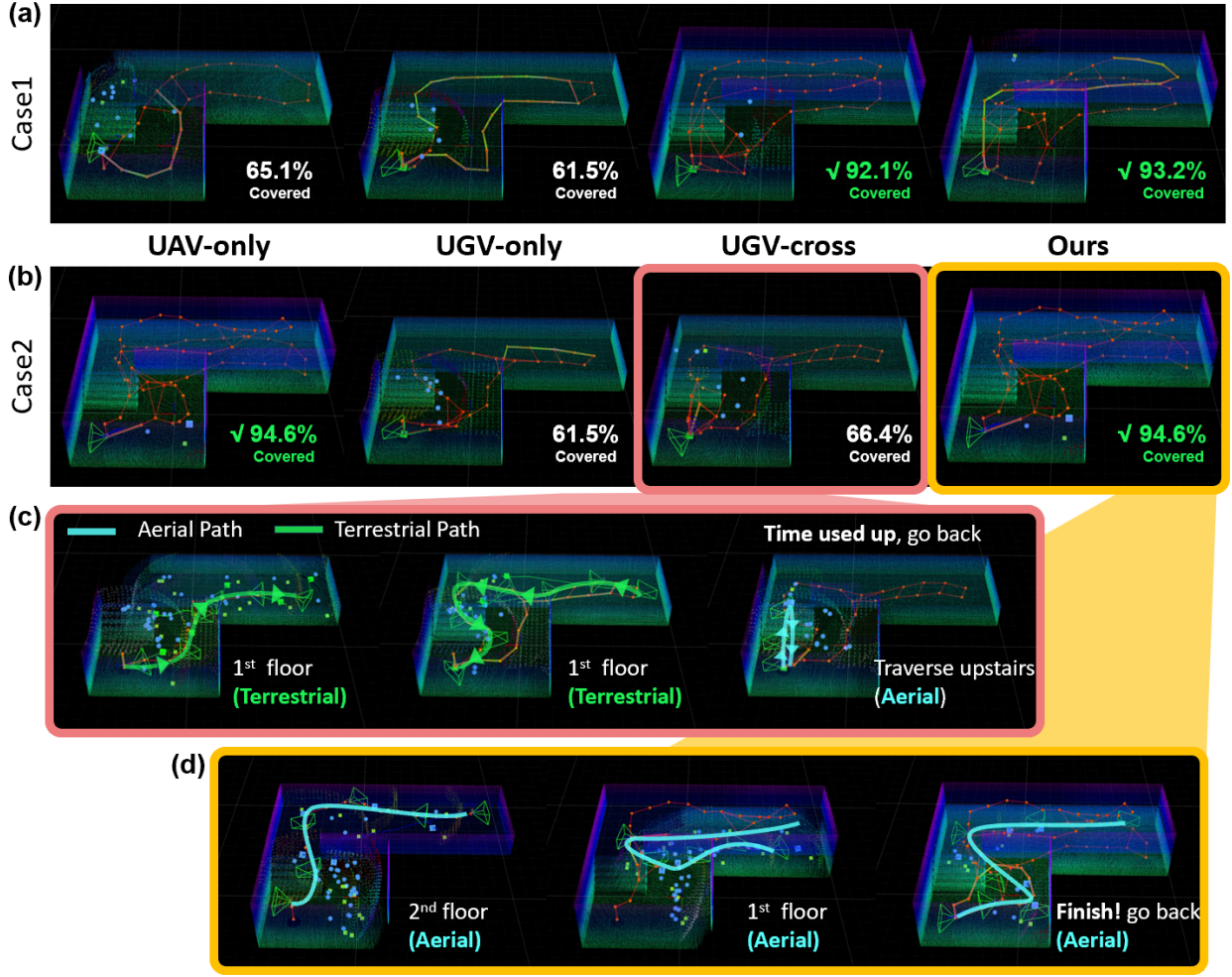

Figure 12: Modality configuration comparisons under two cases. (a). Case 1, where energy budget = 300, time budget = 400. (b). Case 2, where energy budget = 800, time budget = 100. (c). The exploration process of **UGV-cross** system in case 2. The **UGV-cross** variant fails to explore the whole environment when time budget is tightly limited. (d). The exploration process of our system in case 2. Our system can always maintain high coverage of the unknown environment under different resource conditions.

In case 1 (E300, T400) (Fig. 12(a)), where energy budget is tighter, the **UAV-only** system consumes energy rapidly due to sustained flight and is only able to cover 65.1% of the environment before depleting its resources. The **UGV-only** system is unable to access the second floor and therefore can only complete exploration on the ground level. Both the **UGV-cross** and our system are able to almostly complete the full exploration task in this scenario. Note that their energy usage in the table does not reach the maximum budget, this is because one remaining frontier cluster is deemed unreachable: the estimated energy was insufficient to cover that cluster and return to the departure station. Therefore, the robot opted to return directly, leaving that cluster unexplored.

In case 2 (E800, T100) (Fig. 12(b)), where time budget is tighter, the **UAV-only** system completes full exploration within the time budget. The **UGV-only** system again

fails to reach the second floor and is limited to covering the first level. The **UGV-cross** system switches to aerial modality only when the next viewpoint is unreachable in terrestrial modality. As a result, it completes exploration on the first floor, but runs out of time while attempting to ascend via aerial modality, and has to return before reaching the second level. Its coverage is therefore limited to 66.4%. The entire **UGV-cross**'s exploration process is visualized in Fig. 12(c). In contrast, our system anticipates the time constraint and proactively switches to aerial modality, completing the full exploration.

In summary, our system can dynamically adapt to varying energy and time budgets. The BM-MCTS-based planner estimates future resource consumption and evaluates nearby viewpoints with their respective modality and position, enabling flexible and strategic switching to maximize environment coverage under constraints.

## 8.3 Resource-aware Decision-making Algorithm Comparison

We compare our BM-MCTS method with NBV-based and TSP-based method. The NBV-based method adopts a greedy policy by selecting the viewpoint with the highest immediate gain, while the TSP-based method computes a global tour over all viewpoints to minimize total cost.

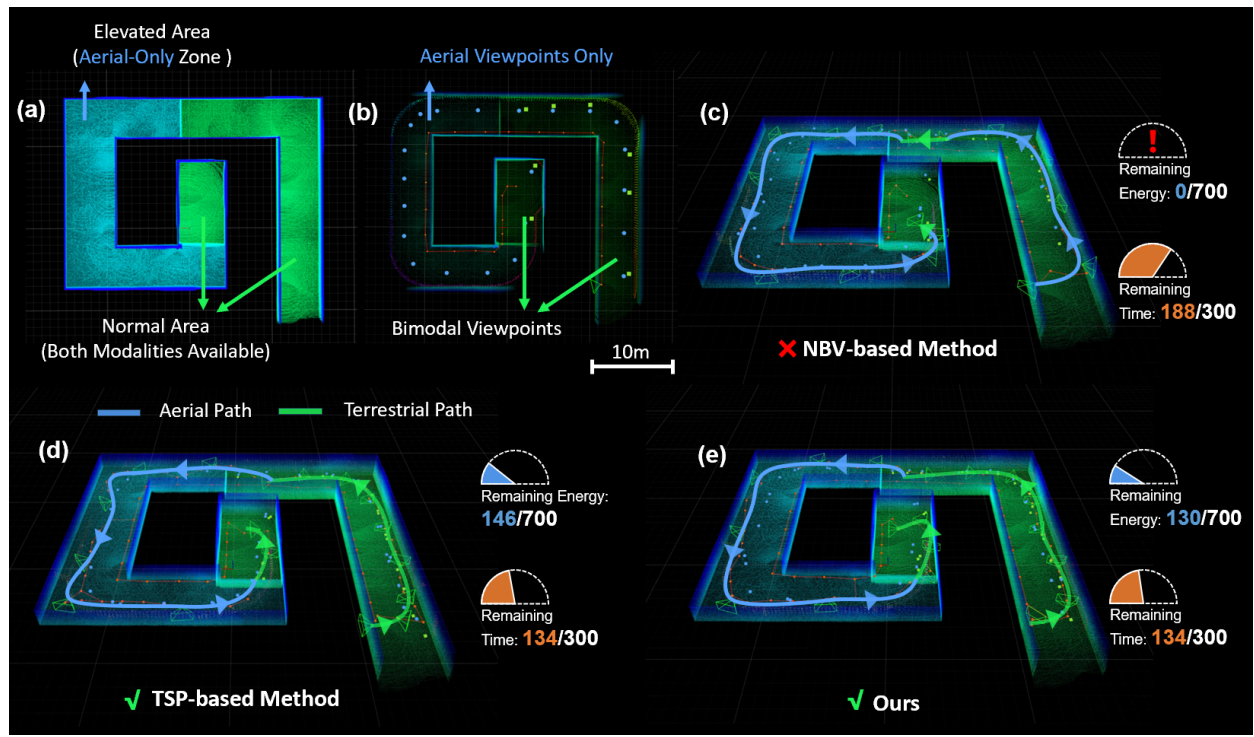

Figure 13: Resource-aware decision-making algorithm comparisons (a). The simulation environment. The blue elevated area represents a region that is untraversable by the terrestrial modality. (b). A half-explored environment that explored by a fix trajectory. (c). The exploration result of the NBV-based method. (d). The exploration result of the TSP-based method. (e). The exploration result of our method.

For the algorithm-level comparison, we retain the same bimodal viewpoint generation and motion planning components, and replace only the BM-MCTS module with either the NBV-based method or the TSP-based method.

For the NBV-based method, we directly evaluate all candidate viewpoints and select the one with the minimum score for execution. To ensure consistency with the scoring mechanism used in the BM-MCTS selection process, we define the score of a viewpoint $\mathcal{P}$ based on three components: (1) $IG(\mathcal{P})$, the information gain, measured by the number of frontier cells visible from $\mathcal{P}$; (2) $\kappa_{E_r}(\mathcal{P})$, a penalty term based on the remaining energy after visiting $\mathcal{P}$; and (3) $\kappa_{T_r}(\mathcal{P})$, a penalty term based on the remaining time after visiting $\mathcal{P}$. The overall score function is defined identically to Equation (6b) in the revised main paper:

$$G(\mathcal{P}) = -N(IG(\mathcal{P})) + \kappa_{E_r}(\mathcal{P}) + \kappa_{T_r}(\mathcal{P}), \tag{1}$$

where $N(x)$ linearly normalizes $x$ to the range $[\epsilon, 1]$, and $\epsilon = 0.05$ denotes the minimum normalized value. The energy and time penalty terms are computed using exponential functions: $\kappa_{E_r}(\mathcal{P}) = exp(-a_1 \cdot E_r(\mathcal{P})/E_{all} + b_1)$, $\kappa_{T_r}(\mathcal{P}) = exp(-a_2 \cdot T_r(\mathcal{P})/T_{all} + b_2)$, where $E_{all}$ and $T_{all}$ denote the total energy and time budgets, respectively, and $E_r(\mathcal{P})$, $T_r(\mathcal{P})$ represent the remaining energy and time after reaching $\mathcal{P}$ from the current pose. The hyperparameters $a_1$, $a_2$, $b_1$, and $b_2$ are set identically to those used in the BM-MCTS algorithm.

For the TSP-based method, we adapt the standard approach to accommodate the bimodal exploration context. For each frontier cluster, we consider two alternative strategies: visiting viewpoints sampled under the aerial modality or under the hybrid (terrestrial-aerial) modality, as described in Sec. IV.A. Therefore, for $n$ frontier clusters, there exist $2^n$ possible modality assignments, and we solve the TSP for each combination to find the optimal visiting sequence. For each candidate sequence $\boldsymbol{\mathcal{P}} = \{\mathcal{P}_1, ..., \mathcal{P}_n\}$, where each $\mathcal{P}_i$ corresponds to a selected viewpoint in a frontier cluster, we compute its score using the same formulation:

$$G(\boldsymbol{\mathcal{P}}) = -N(IG_{\mathrm{d}}(\boldsymbol{\mathcal{P}})) + \kappa_{E_r}(\boldsymbol{\mathcal{P}}) + \kappa_{T_r}(\boldsymbol{\mathcal{P}}), \tag{2}$$

where $IG_{\mathrm{d}}(\boldsymbol{\mathcal{P}})$ denotes the discounted average information gain of the entire viewpoint sequence $\boldsymbol{\mathcal{P}}$ (to be consistent to the IG-related item in Equation (5a) in the revised main paper), computed as:

$$IG_{\mathrm{d}}(\boldsymbol{\mathcal{P}}) = \frac{IG(\mathcal{P}_1) + \gamma_{\mathrm{IG}} \; IG(\mathcal{P}_2) + \cdots + \gamma_{\mathrm{IG}}^{n-1} \; IG(\mathcal{P}_n)}{1 + \gamma_{\mathrm{IG}} + \cdots + \gamma_{\mathrm{IG}}^{n-1}}, \tag{3}$$

where $\gamma_{\mathrm{IG}}$ is the discount factor (same as $\gamma_{\mathrm{IG}}$ in the revised main paper Alg. 2). And the energy/time terms $\kappa_{E_r}(\boldsymbol{\mathcal{P}})$ and $\kappa_{T_r}(\boldsymbol{\mathcal{P}})$ represent the remaining resources after completing the full trajectory and returning to the departure station. The sequence with the minimum $G(\boldsymbol{\mathcal{P}})$ is selected for execution. While the TSP-based method provides a globally optimal solution, the BM-MCTS algorithm relies on its simulation phase to estimate energy and time costs for portions of the search tree that have not yet been expanded.

We compare the three algorithms in the simulation shown in Fig. 13(a). To ensure a fair comparison under stochastic decision-making, we employ a simulation environment with a single-path topological structure to ensure consistent spatial topology in exploration across all methods. In the elevated area, only aerial viewpoint can be generated, and in the

normal area, viewpoints of both modalities can be generated. We first make the environment half-explored with a fixed trajectory (Fig.13(b)), and pre-generate some frontier clusters and viewpoints. Then, we start the exploration with energy budget = 700 and time budget = 300.

The results indicate that the NBV-based method is inherently shortsighted. Since it only considers the immediate next viewpoint, it fails to proactively switch to the terrestrial modality early enough in anticipation of the later aerial-only zone. As a result, the robot depletes its energy budget before completing the full exploration. In contrast, both the TSP-based method and our proposed BM-MCTS approach perform long-horizon planning. They are able to foresee the upcoming aerial-only region and conserve energy in advance by utilizing the more energy-efficient terrestrial modality earlier in the mission.

However, obtaining a globally optimal solution with the TSP-based method comes at a high computational cost. As illustrated in Fig. 14, the computation time increases rapidly as the number of frontier clusters grows.

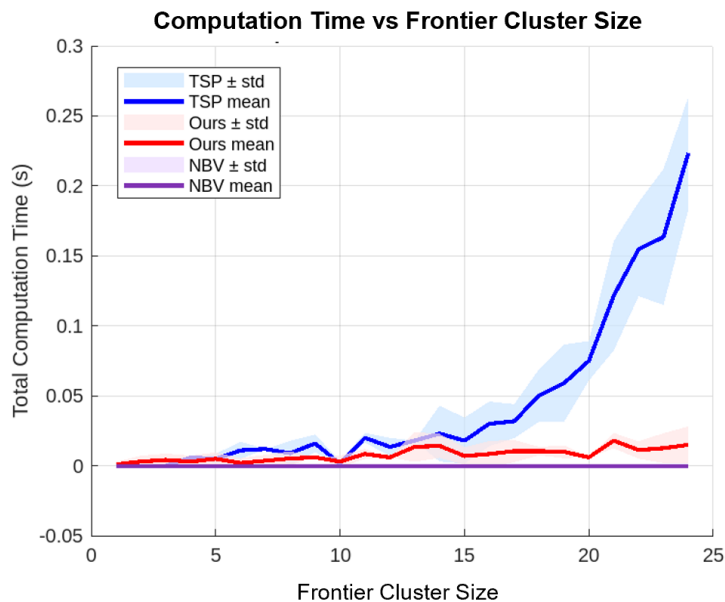

Figure 14: Computation time comparison among NBV-based, TSP-based and our algorithms.

# 9 Details of Problem Formulation

## 9.1 Hyperparameter Selection of the Penalty Function

Recall our motivation in the context of search and rescue operations. Disasters such as earthquakes and fires can alter the structure of environments, posing significant risks to human responders. Deploying robots mitigates these risks and enables exploration and data collection in areas that are otherwise inaccessible to humans.

In many disaster scenarios, communication infrastructure is often compromised. As a result, ensuring data retrieval by having robots return to the departure station becomes the

most reliable solution. Furthermore, the primary objective of exploration in such environments is to locate and rescue trapped individuals. Timely access to firsthand situational data is crucial for rescue teams, often within a limited timeframe [7], which introduces constraints on the expected exploration duration.

Given this practical context, ensuring sufficient remaining energy to return to the departure station is of utmost importance. In contrast, time constraints are generally softer—minor overruns are typically tolerable. As such, we design the penalty functions accordingly: the energy-related penalty is sharper, while the time-related penalty is flatter and only becomes influential when sufficient energy remains. Therefore, we apply a sharper penalty function for remaining energy and a flatter one for remaining time. Accordingly, we set the parameters as $a_1 = b_1 - log(0.3)$, $b_1 = log(10)$, $a_2 = b_2 - log(0.7)$, and $b_2 = log(3)$, as shown in Fig. 15.

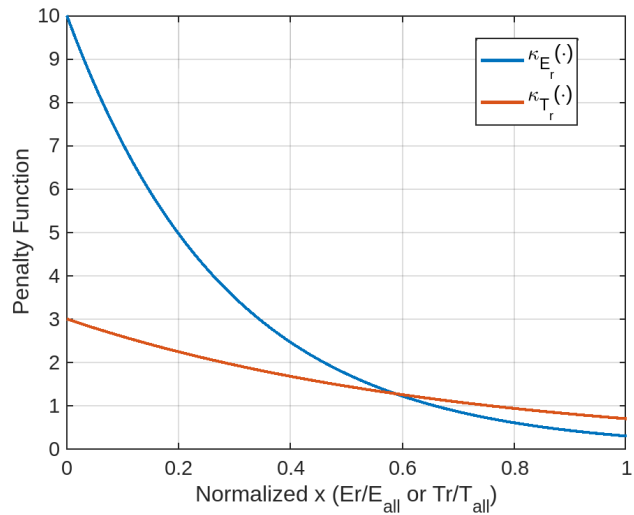

Figure 15: Penalty functions of reamining energy and time.

Such penalty functions also drive the robot to cope with environmental uncertainty by maintaining a safety margin.

## 9.2   IG Normalization

For IG, we apply a normalization function $N(\cdot)$ to obtain $N(R_p(v_i))$, where $R_p$ denotes the IG-related process gain. $N(\cdot)$ linearly maps the value to the range $[\epsilon, 1]$, where $\epsilon = 0.05$ denotes the minimum normalized score.

For a node $v$, this normalization is performed over all $v$'s child nodes $\{v_i \mid i = 0, \ldots, n\}$, based on their corresponding IG-related process gain $R_p(v_i)$. The normalization function is defined as:

$$N(R_p(v_i)) = \begin{cases} 0.5, & \text{if } |R_{p,\max} - R_{p,\min}| \leq \delta \\ \epsilon + (1-\epsilon) \cdot \dfrac{R_p(v_i) - R_{p,\min}}{R_{p,\max} - R_{p,\min}}, & \text{otherwise} \end{cases}$$

where $R_{p,\min}$ and $R_{p,\max}$ are the minimum and maximum values among the child nodes, $\delta$ is a small threshold that defines a dead zone for normalization, preventing numerical instability when the difference between $R_{p,\max}$ and $R_{p,\min}$ is very small.

We adopt this normalization to ensure that the scores of all nodes are mapped to a comparable, dimensionless scale. This allows the IG-related term to be balanced meaningfully against the energy/time penalties. Note that in our bimodal viewpoint generation stage, each viewpoint is guaranteed to have a positive IG (i.e., it observes nonzero frontier voxel), so we use a small floor value like $\epsilon = 0.05$ to represent $R_{p,\min}$ after normalization.

# 10    Details of Hardware and FoV

In Fig. 16, we provide a photograph of the TABV platform that illustrates its hardware components, including the sensor placement and actuation mechanisms.

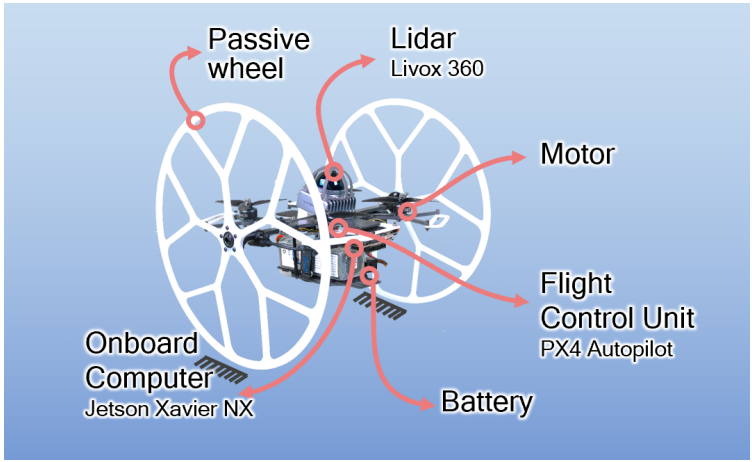

Figure 16:   The hardware of the TABV.

Although the sensor remains the same, the effective FoV differs between modalities due to its vertically symmetric mounting on the TABV. This design balances two competing needs: i) In terrestrial modality, more downward visibility is needed to percept the ground, especially the nearby ground, which is critical for motion planning. ii) In general, upward visibility is essential for perceiving the surrounding environment. As a result of this compromise, the effective perceptual FoV in terrestrial modality is approximately half that in aerial modality, as shown in Fig. 17(a).

Note that our method is compatible with any sensor with a limited FoV, and the FoV is explicitly passed as a parameter to the planner. The viewpoint generation process dynamically adapts to the specified FoV of the sensor. Specifically, the FoV parameter will be used in the coverage calculation in Alg. 1.

In the simulation experiments presented in Sec. VII.A (Fig. 7 of the main paper), we use a depth sensor with a horizontal FoV of 90°, vertical FoV of 60°, and a sensing range of 3.5 meters. For the large-scale exploration task in a two-story house in Gazebo (Fig. 8), the TABV is equipped with a 360° LiDAR sensor with a vertical FoV ranging from −30° to 30° and a maximum sensing range of 4.5 meters. In the real-world experiments, we mount

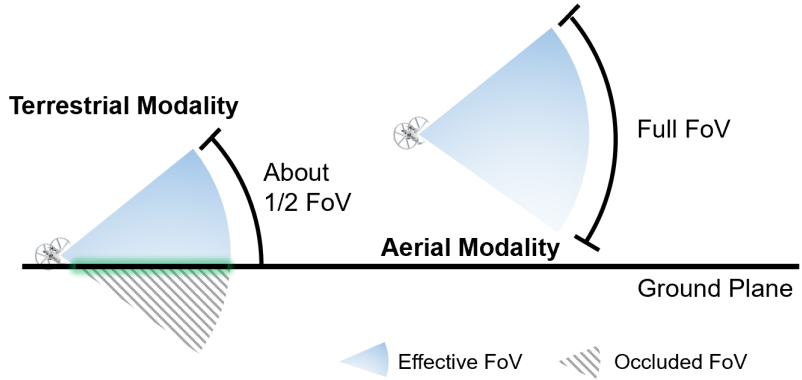

Figure 17: The effective FoV of different modalities.

a Livox Mid-360 LiDAR with a tilted configuration to obtain a vertically symmetric FoV. Due to this mounting scheme, the LiDAR effectively observes only the front-facing 180°, leading to a usable FoV of approximately 60° × 180°. Additionally, to facilitate real-world validation and account for the relatively sparse point cloud density of the Livox Mid-360 at longer distances, we set the effective sensing range to 5 meters to ensure reliable perception.

## 11    Limitation and Future Work

i) **Locally myopic resource estimation.** The estimation of required energy and time is based solely on the currently observed environment. This local perspective may lead to shortsighted planning decisions. In future work, we plan to incorporate environment prediction modules as priors to enable more informed and globally-aware resource allocation during exploration.

ii) **High System complexity.** Our proposed system integrates multiple components: viewpoint generation, topological graph construction, modality-aware decision-making, bimodal mapping, localization, bimodal motion planning and bimodal MPC controller, which makes the overall pipeline relatively complex and harder to maintain. We plan to explore more end-to-end learning frameworks, particularly for control and decision-making, to reduce system complexity and minimize the gap between modules.

iii) **Simplified terrain analysis.** The current method determines ground traversability based on extracted ground surfaces from a grid map. However, this process is relatively simplistic, and its accuracy depends heavily on the resolution of the grid map. In future work, we plan to maintain a more detailed elevation map and perform traversability analysis based on terrain structure to improve the reliability of ground path planning.



\end{document}